\documentclass{article}

\usepackage[preprint]{neurips_2026}
\usepackage[utf8]{inputenc}
\usepackage[T1]{fontenc}
\usepackage{float}
\usepackage{placeins}
\usepackage[hidelinks,breaklinks=true]{hyperref}
\usepackage{url}
\usepackage{booktabs}
\usepackage{amsfonts}
\usepackage{nicefrac}
\usepackage{microtype}
\usepackage[table]{xcolor}
\usepackage{tabularx,array,graphicx,makecell}
\usepackage{tikz}
\usetikzlibrary{arrows.meta,positioning,calc,fit}

\newcommand{\systemName}{{\textsc{WildTrace}}}
\newcommand{\releaseRepo}{\href{https://huggingface.co/datasets/CinderD/wildtrace}{\texttt{CinderD/wildtrace}}}

\newcolumntype{Y}{>{\raggedright\arraybackslash}X}
\newcolumntype{L}[1]{>{\raggedright\arraybackslash}p{#1}}

\usepackage{amsmath} 
\title{\systemName{}: Benchmarking Natural Evidence Trails in Long-Context Reasoning}

\author{
  $\textbf{Zixin Chen}^{1,2}\thanks{Work done during internship at Qwen Team} \;\; \textbf{Peng Liu}^2 \;\; \textbf{Haobo Li}^1 \;\; \textbf{Rui Sheng}^1 \;\; \textbf{Jianhong Tu}^2$ \\
  $\textbf{Xiaodong Deng}^2 \;\; \textbf{Fei Huang}^2 \;\; \textbf{Kashun Shum}^{1,2} \;\; \textbf{Dayiheng Liu}^2 \;\; \textbf{Huamin Qu}^1$ \\
  $^1$ Hong Kong University of Science and Technology \\
  $^2$ Qwen Team, Alibaba Group \\
  \texttt{zchendf@connect.ust.hk} \\
}

\begin{document}

\maketitle

\begin{abstract}
Answering complex questions over long documents frequently requires integrating evidence that the source itself disperses naturally across distant passages. In an incident report, the operating condition, design flaw, and missed safety check that jointly explain a disaster may appear dozens of sections apart; in a novel, a character's true motive may surface only through scenes far removed from the moment it becomes relevant. This source-internal evidence integration is central to real-world long-document analysis, yet existing benchmarks largely sidestep it. Needle probes, planted facts, and reverse-engineered multi-hop chains embed evidence that may differ from the host text in distribution, placement, or register, making it unclear whether strong performance reflects genuine source reasoning or distributional artifacts. We introduce \systemName{}, a benchmark of 481 tasks over 214 naturally occurring long-form sources such as technical incident reports and lesser-known literary narratives, where all evidence trails arise from the document's own causal, temporal, and narrative logic. Drawing on Pearl's causal hierarchy and prior multi-hop reasoning typologies, we define seven source-internal evidence geometries that characterize the distinct relational demands of analytical reading in long documents. A source-first construction pipeline mines candidate trails from document structure before writing questions; each item then undergoes multi-stage validation covering clue necessity, answer groundedness, rubric fidelity, contamination resistance and answerability. Across 18 frontier systems evaluated under evidence-withheld conditions, the strongest reaches 75.3\%, with pronounced weaknesses on reasoning-intensive geometries such as counterfactual branching and causal attribution. As models are increasingly entrusted with real-world high-stakes analytical tasks, this gap between accessing information and reasoning over naturally dispersed evidence emerges as a defining challenge for the next stage of long-context research.
\end{abstract}

\section{Introduction}

Long-document reasoning is not merely retrieval over more tokens. For a given analytic question, a model must decide which ordinary source fragments are load-bearing evidence, which same-register details are distractors, and which causal, temporal, comparative, abductive, or counterfactual dependency makes the selected fragments jointly warrant an answer. In literary analysis, a character's motive may not reside in the scene where the decisive action occurs; it may be warranted only by linking an early promise, a later contradiction, and a delayed consequence that changes how earlier scenes should be read. Structured technical documents pose the same burden in a more procedural register. Even when a report contains findings, headings, tables, and cross-references, a follow-up question may require linking an operating condition, a design constraint, and a review lapse whose dependency is not packaged as the answer to that query. The relevant passages may all be locatable, but the task is to recover why they jointly support the conclusion.

Many existing evaluations measure important adjacent capabilities, but often control the evidence environment in ways that reduce the burden of evidence discovery and relation recovery. Needle-style and controlled long-context probes measure access, position sensitivity, aggregation, and robustness, yet the relevant facts are inserted or otherwise benchmark-controlled \citep{liu2024lost,hsieh2024ruler,kuratov2024babilong,li2024needlebench}. Pre-selected-passage and retrieval-grounded settings test whether a model can use supplied support, but the task interface has already decided which passages matter~\citep{bai2024longbench,yen2024helmet}. Compositional multi-hop datasets test bridge following, comparison, and shortcut-resistant reasoning, but their
chains are often assembled or reverse-engineered around target answers rather than recovered from a raw source's own narrative, procedural, and logical structure~\citep{yang2018hotpotqa,ho2020constructing,trivedi2022musique}.

This tradeoff is not merely philosophical. Prior analyses show that nominally multi-hop questions can be answerable from single-hop evidence or dataset artifacts, while long-context performance can depend strongly on where evidence appears and how literally the query matches the target span
\citep{min2019compositional,chen2019understanding,gururangan2018annotation,
mccoy2019right,liu2024lost,modarressi2025nolima}. Inserted or localized evidence can therefore introduce construction cues---position, lexical overlap, evidence density, or stylistic mismatch with surrounding text---that need not be present when the relevant fragments are naturally authored within the source. In downstream analytical use, support is rarely pre-marked for the query: an analyst gives the model a source and asks for a conclusion that can be warranted from it. The missing evaluation boundary is therefore not long context or multi-hop reasoning in isolation, but full-document, evidence-withheld reasoning over source-induced evidence trails: deciding what counts as evidence, preserving why
it counts, and reasoning over the dependency that makes it support the answer.

We introduce \systemName{}, a benchmark for multi-hop reasoning over \emph{natural evidence trails} in long contexts. A natural evidence trail is a set of source-internal clues and relations that jointly warrant an answer, with both the clue layout and the inferential dependency induced by the document itself. \systemName{} is therefore built source-first rather than question-first. Starting from 214 public long-context sources, including technical incident reports and lesser-known literary narratives, we segment each source into overlapping windows, convert local context into evidence records---entities, events, attributes, temporal markers, quotations, local relations, and short summaries---and link these records into a typed source-internal graph. An edge is added only when the surrounding source text supports the relation between the linked facts. Geometry-specific queries then mine naturally occurring trails for seven evidence geometries: forward chains, intersections, comparisons, temporal reconstructions, causal fan-in, abductive explanations, and counterfactual branches. Only after such a trail has been grounded do we write the public question, reference answer, key facts, and criterion-level rubric. Across construction, this source-first pipeline instantiated 3,506 validation-tracked candidate trail--question artifacts.

Candidate mining is high-recall by design: a trail becomes a benchmark item only after verification. Natural long-context questions can look multi-hop while still
being answerable from a single local clue, from public familiarity, from an unsupported fact, or from a reasoning pattern different from the assigned geometry. We therefore organize promotion around four threats to benchmark fidelity: source grounding, multi-hop necessity, shortcut resistance, and answerability. Grounding checks verify that answers, key facts, evidence spans, and rubric criteria are supported by the source and reward answer-critical relations rather than decorative details. Necessity and shortcut checks use
leave-one-out and single-clue ablations to reject items where a proper subset of the evidence already yields the unique answer, while geometry-specific review rejects collapsed relations. No-document probes filter contamination and public-knowledge shortcuts. Evidence-conditioned checks verify that the hidden trail is
sufficient when supplied, and human-led plus external audits catch unsupported facts, redundant clues, ambiguous answers, and geometry mismatches. After these
gates, the public release\footnote{Public dataset release: \releaseRepo{}.} retains 481 accepted tasks over 214 long-form sources, a 13.7\% survival rate.

We evaluate 18 frontier systems under full-document, evidence-withheld prompts using three non-contestant judges. The strongest evaluated system reaches 75.3\%, leaving substantial headroom even for frontier long-context models. The main result is not a top-1 ranking but
a diagnostic account of where source-internal reasoning breaks. Models often produce locally plausible answers while failing the evidence relation: they name the right entity but drop a disambiguating constraint, recover one local story but miss the critical comparison, list plausible causal ingredients without the dependency that makes them explanatory, or follow the factual path while losing the counterfactual branch. Performance also varies by source family, context tier, and evidence geometry. These patterns suggest that the
bottleneck is not only how much context a model can ingest, but whether it can turn natural evidence trails into warranted answers by selecting the right source fragments and preserving the relation among them.

Our contributions are threefold:
\begin{itemize}
    \item \textbf{A source-internal formulation of long-context reasoning.}
    We define natural evidence trails as source-induced configurations of
    dispersed evidence that jointly warrant an answer, and organize them into
    seven evidence geometries capturing distinct relational demands.

    \item \textbf{A source-first, validity-gated benchmark release.}
    \systemName{} contains 481 tasks over 214 long-form source files, distilled
    from 3,506 validation-tracked candidates. Evidence structures are identified
    before question writing, and each released item undergoes multi-stage checks
    of source support, evidence necessity, answerability, rubric fidelity, and
    geometry consistency.

    \item \textbf{A diagnostic evaluation of frontier systems.}
    We evaluate 18 systems under a full-document, evidence-withheld protocol
    with criterion-level multi-judge scoring. The results reveal substantial
    remaining headroom, systematic performance changes with document scale,
    and distinct reasoning profiles across evidence geometries.
\end{itemize}

\section{\systemName{} Benchmark and Evaluation Design}
\label{sec:benchmark-construction}

\subsection{A Running Example: Locating Evidence Is Not Enough}
\label{sec:running-example}

We start with a task case as it shows the full capability that \systemName{} evaluates: precise evidence localization, source-grounded support, and reasoning over the dependency that makes dispersed clues jointly answer the question. Figure~\ref{fig:main-case-examples} shows a locked L6 comparative item from a roughly 950K-token literary source. At test time, the model receives the full novel and a public question asking it to compare two distant episodes in which missing communication makes an innocent character appear morally guilty: Jack Tanerton's sailing on the \emph{Rose of Delhi} and Mamie Lee's contested marriage to James West. The answer-critical evidence is not localized in a single passage. Two early windows at 9.2--9.3\% establish Jack's apparent fault
and Aunt Dean's suppression of the letters that would have explained him; three much later windows at 80.2--81.6\% establish Mamie's apparent fault, the documentary reversal of that judgment, and the failed wartime contact path.

The model must do more than retrieve these windows or repeat their shared topic. The full-credit answer is a mechanism-level comparison. Jack's apparent disobedience is manufactured by deliberate suppression: Aunt Dean prevents the relevant letters from reaching their recipients, making Jack's departure appear willful. Mamie's apparent guilt is undone differently: documentary proof of marriage and James West's failed contact path explain why silence was not abandonment. A response that recovers only one episode, or says only that both cases involve missing communication, is relevant but incomplete. It identifies the shared surface frame, but misses the mechanism-level comparison: Jack's false guilt is created by deliberate suppression of letters, whereas Mamie's is resolved through documentary proof and failed wartime contact.

The construction artifact turns this mechanism-level requirement into a checkable benchmark item. It stores the five load-bearing source windows, the grounded spans within them, the typed comparative relation, a reference answer, criterion-level rubric, and leave-one-out checks. Removing Jack's windows removes the deliberate-suppression mechanism; removing Mamie's windows removes the documentary reversal and failed-contact mechanism; removing the contrast leaves two local plot summaries rather than the requested comparison. These artifacts are used for validation and scoring, but they are withheld from the evaluated model. The model sees only the full source and the public question, and earns full credit only if it locates the right evidence and preserves the cross-case relation that makes the answer warranted.

\begin{figure}[H]
\centering
\definecolor{wtCaseHeader}{HTML}{EAF2F8}
\definecolor{wtCaseBlue}{HTML}{245A8D}
\definecolor{wtCaseBlueLight}{HTML}{EEF5FB}
\definecolor{wtCaseGold}{HTML}{A76A18}
\definecolor{wtCaseGoldLight}{HTML}{FFF4E6}
\definecolor{wtCaseGreen}{HTML}{2E7D5B}
\definecolor{wtCaseGreenLight}{HTML}{EAF6EF}
\definecolor{wtCaseRed}{HTML}{A33A3A}
\definecolor{wtCaseRedLight}{HTML}{FDEEEE}
\definecolor{wtGemini}{HTML}{4285F4}
\definecolor{wtOpenAI}{HTML}{111827}
\definecolor{wtQwen}{HTML}{6D5BD0}
\definecolor{wtMini}{HTML}{D66E2A}
\newcommand{\caseTag}[2]{\tikz[baseline=-0.55ex]\node[rounded corners=2pt, draw=#1!70!black, fill=#1!10, inner xsep=2.6pt, inner ysep=1.1pt, font=\sffamily\scriptsize\bfseries, text=#1!70!black]{#2};}
\newcommand{\clueChip}[2]{\tikz[baseline=-0.58ex]\node[rounded corners=2pt, draw=#1!50!black, fill=#1!7, inner xsep=2.8pt, inner ysep=1.0pt, font=\sffamily\scriptsize\bfseries, text=#1!70!black]{#2};}
\newcommand{\clueLine}[4]{\clueChip{#1}{#2}\hspace{0.32em}\textbf{#3:} #4}
\newcommand{\blockgap}{\addlinespace[2.0pt]}
\newcommand{\miniCell}[2]{\textbf{#1:} #2}
\newcommand{\modelChip}[4]{%
\tikz[baseline=-0.58ex]{
  \node[circle, fill=#1, text=white, inner sep=0pt, minimum size=1.72ex, font=\sffamily\tiny\bfseries] (icon) at (0,0) {#2};
  \node[anchor=west, rounded corners=2pt, draw=#1!50!black, fill=#1!7, inner xsep=2.6pt, inner ysep=1.1pt, font=\sffamily\scriptsize\bfseries, text=#1!65!black] at (0.78ex,0) {#3 \textcolor{black!55}{#4}};
}}
\newcommand{\modelLine}[5]{\modelChip{#1}{#2}{#3}{#4}\quad #5}
\footnotesize
\setlength{\tabcolsep}{4.2pt}
\renewcommand{\arraystretch}{1.18}
\begin{tabularx}{\linewidth}{@{}L{0.17\linewidth}Y@{}}
\toprule
\rowcolor{wtCaseHeader}
\multicolumn{2}{@{}p{\linewidth}@{}}{\textbf{Case Card: Comparative Evidence Trail}\par
\textbf{Task ID:} \texttt{johnny\_ludlow\_series\_1\_2\_early3\_b5c-comp-r238-0001}} \\
\midrule
\caseTag{wtCaseBlue}{TASK} &
Compare two distant episodes in which missing communication makes an innocent character look morally guilty: Jack Tanerton sailing on the \emph{Rose of Delhi}, and Mamie Lee's contested marriage to James West. \\
\blockgap
\rowcolor{wtCaseGoldLight}
\caseTag{wtCaseGold}{GOLD} &
\miniCell{Jack}{Aunt Dean burns both Jack's farewell and the Rector's no-sail letter, so Jack is wrongly judged to have defied a refusal he never received.}\par
\miniCell{Mamie}{James's Bristol certificate proves the marriage, and his letters to Ireland plus battlefield injury explain why his silence was not abandonment.}\par
\miniCell{Contrast}{both episodes reverse moral blame through communication evidence, but the mechanisms differ: deliberate suppression versus missing proof and failed wartime contact.} \\
\blockgap
\rowcolor{wtCaseBlueLight}
\caseTag{wtCaseBlue}{ANCHORS} &
\clueLine{wtCaseRed}{C1}{@9.2\%, hidden refusal}{Aunt Dean burns Jack's farewell and the Rector's no-sail letter.}\par
\clueLine{wtCaseGold}{C2}{@9.3\%, moral consequence}{Jack is read as disobedient and nearly disinherited.}\par
\clueLine{wtCaseRed}{C3}{@80.2\%, apparent fault}{Mamie returns with a child but no marriage lines, making her claim suspect.}\par
\clueLine{wtCaseGreen}{C4}{@81.5\%, documentary proof}{James returns with the Bristol certificate proving the marriage.}\par
\clueLine{wtCaseBlue}{C5}{@81.6\%, missing contact}{James's Ireland letters, wounding, and discharge explain the silence.} \\
\blockgap
\rowcolor{wtCaseGoldLight}
\caseTag{wtCaseGold}{ROUTE TEST} &
\textbf{C1} gives deliberate suppression, \textbf{C2} gives its moral effect, \textbf{C3} creates Mamie's apparent guilt, \textbf{C4} overturns it with proof, and \textbf{C5} explains the failed contact; the answer must contrast these mechanisms, not just name missing communication. \\
\blockgap
\rowcolor{wtCaseGreenLight}
\caseTag{wtCaseGreen}{PASS} &
\modelLine{wtGemini}{G}{Gemini-3.1-Pro}{96.7}{\textbf{Mechanism contrast:} Jack's false guilt is engineered by Aunt Dean's intentional interception, while Mamie's false guilt is overturned by James's certificate, misdirected letters, and wartime injury.} \\
\blockgap
\rowcolor{wtCaseRedLight}
\caseTag{wtCaseRed}{FAIL} &
\modelLine{wtOpenAI}{O}{GPT-5.1}{45.7}{\textbf{Missing one side:} recovers parts of the letter-suppression plot but loses the full Mamie reversal, so the comparison is under-supported.}\par
\modelLine{wtQwen}{Q}{Qwen3.7-Plus}{36.7}{\textbf{Collapsed mechanism:} keeps the shared ``missing communication'' theme but merges suppression, marriage proof, and misdirected wartime letters into one generic misunderstanding.}\par
\modelLine{wtMini}{M}{MiniMax-M2.7}{0.0}{\textbf{Context-window limit:} this L6 source is about 950K estimated tokens, exceeding MiniMax's supported context window, so the full prompt was not sent and the item receives no route credit.} \\
\bottomrule
\end{tabularx}
\caption{\textbf{A WILDTRACE task design example.} The public task asks for a comparative evidence trail; the card displays construction artifacts that are hidden during evaluation. \textbf{C1--C5} are the five load-bearing clue windows, and \textbf{@} marks each window's approximate source location. The item tests whether a model can recover both mechanisms across distant passages: Jack's apparent guilt comes from deliberate letter suppression, whereas Mamie's is resolved by documentary proof plus failed wartime contact. \textbf{PASS/FAIL} rows summarize representative route outcomes: high credit preserves this contrast; partial credit names nearby facts but drops or collapses a mechanism; the context-window row shows an endpoint that could not receive the full source and therefore receives no item credit.}
\label{fig:main-case-examples}
\end{figure}

\subsection{Benchmark Object and Release Snapshot}
\label{sec:benchmark-overview}
\label{sec:task-definition}

The running example instantiates the benchmark object that \systemName{} constructs at scale. Each item begins with one raw long-form source and a naturally occurring evidence trail: a set of source spans and typed relations that jointly warrant an answer. The public task exposes only the full source and a question. The construction record retains the hidden trail, primary geometry label, reference answer, criterion-level rubric, and validation outcomes for audit and scoring, but none of these artifacts are shown to the evaluated model.

Formally, a release item is a tuple \(x = (S, q, T, a, R)\). Here \(S\) is the raw long-form source, \(q\) is the public question, \(T = (\text{spans}, \text{edges}, g)\) is the hidden evidence trail with primary geometry \(g\), \(a\) is the reference answer, and \(R\) is the criterion-level rubric. At test time, only \((S, q)\) is visible. The model must therefore recover which passages are evidential and preserve the relation that makes them jointly support an answer; evidence spans, clue counts, graph paths, reference answers, rubrics, and validation notes are withheld.

\begin{figure}[!htbp]
  \centering
  \includegraphics[width=\textwidth]{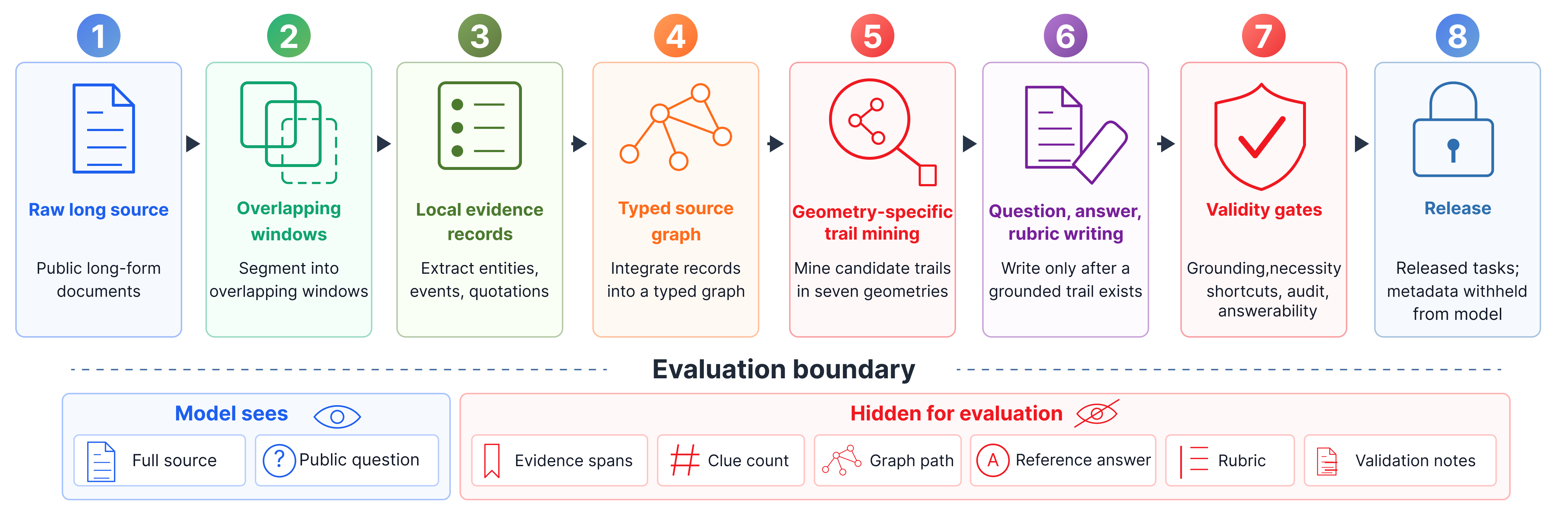}
  \caption{\textbf{\systemName{} construction and evaluation boundary.}
  The pipeline is source-first: long sources are segmented into local records,
  linked into typed source-internal graphs, mined for geometry-specific evidence
  trails, and only then converted into questions, answers, and rubrics. At
  evaluation time, models see only the full source and public question; evidence
  spans, clue counts, graph paths, reference answers, rubrics, and validation
  notes remain hidden and are used only for audit and scoring.}
  \label{fig:benchmark-overview}
\end{figure}

The finalized release contains 481 locked tasks over 214 long-form sources, selected from 3,506 validation-tracked candidate trail--question artifacts. It covers seven evidence geometries, meets the powered L0--L6 geometry-by-tier sample floor, and includes three L7 stress probes. Figure~\ref{fig:benchmark-overview} summarizes the construction and evaluation boundary; Appendix Figure~\ref{fig:release-statistics} and Appendix~\ref{app:source-geometry-artifacts} give the release-composition snapshot.

\subsection{Source Regimes and Evidence Geometries}

The release combines three public source regimes because source-internal evidence takes different forms across document types. Technical incident reports stress causal, procedural, timeline, measurement, and corrective-action evidence. English literary narratives stress motive, delayed revelation, perspective shift, discourse order, and implicit explanation. Chinese literary sources provide an initial Chinese literature slice and diversify the release beyond English without treating \systemName{} as a comprehensive multilingual benchmark.

Context length is stratified for balance and analysis, not treated as the sole definition of difficulty. We use half-octave full-document tiers: \textsf{L0--L6} are the powered release tiers and every geometry-by-tier cell meets the final sample floor, while \textsf{L7} contains only three \(>\)1M-token stress probes. Exact tier boundaries and counts are reported in Appendix Table~\ref{tab:context-tier-coverage}.

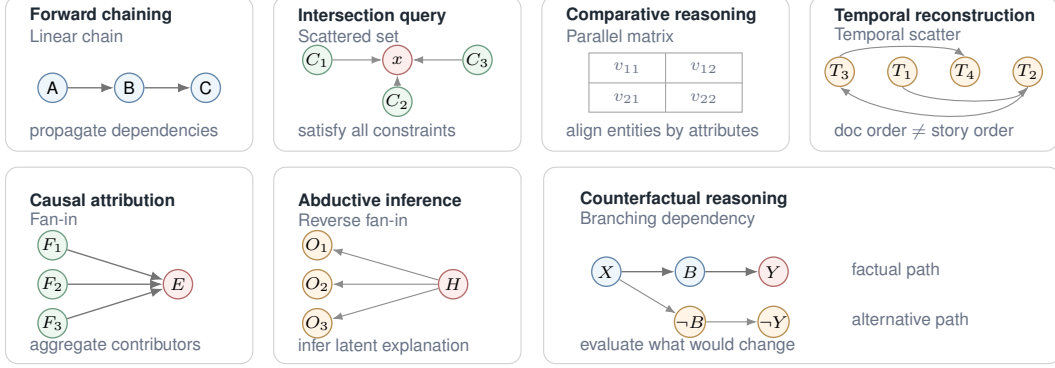
\begin{figure}[t]
\centering
\definecolor{wtInk}{HTML}{1F2933}
\definecolor{wtMuted}{HTML}{697386}
\definecolor{wtBlue}{HTML}{2F5F8F}
\definecolor{wtBlueLight}{HTML}{EEF5FB}
\definecolor{wtGold}{HTML}{A06A1A}
\definecolor{wtGoldLight}{HTML}{FFF4E6}
\definecolor{wtGreen}{HTML}{3E7C59}
\definecolor{wtGreenLight}{HTML}{EDF7F0}
\definecolor{wtRed}{HTML}{9B3A3A}
\definecolor{wtRedLight}{HTML}{FDEEEE}

\resizebox{\linewidth}{!}{%
\begin{tikzpicture}[
  font=\sffamily,
  card/.style={draw=black!18, rounded corners=5pt, fill=white, line width=.45pt, minimum width=3.55cm, minimum height=2.22cm},
  bottomcard/.style={card, minimum height=2.82cm},
  title/.style={font=\sffamily\bfseries\scriptsize, text=wtInk, align=left},
  sub/.style={font=\sffamily\scriptsize, text=wtMuted, align=left},
  nodep/.style={circle, draw=wtBlue!75, fill=wtBlueLight, line width=.55pt, minimum size=4.2mm, inner sep=0pt, font=\sffamily\scriptsize},
  nodeg/.style={circle, draw=wtGreen!75, fill=wtGreenLight, line width=.55pt, minimum size=4.2mm, inner sep=0pt, font=\sffamily\scriptsize},
  nodeo/.style={circle, draw=wtGold!75, fill=wtGoldLight, line width=.55pt, minimum size=4.2mm, inner sep=0pt, font=\sffamily\scriptsize},
  noder/.style={circle, draw=wtRed!75, fill=wtRedLight, line width=.55pt, minimum size=4.2mm, inner sep=0pt, font=\sffamily\scriptsize},
  arr/.style={-{Latex[length=1.8mm]}, draw=black!55, line width=.55pt},
  thinarr/.style={-{Latex[length=1.5mm]}, draw=black!45, line width=.45pt},
  dashline/.style={draw=black!35, dashed, line width=.45pt}
]

\node[card] at (0,2.75) {};
\node[title, anchor=west] at (-1.55,3.58) {Forward chaining};
\node[sub, anchor=west] at (-1.55,3.30) {Linear chain};
\node[nodep] (fa) at (-1.10,2.55) {A};
\node[nodep] (fb) at (0,2.55) {B};
\node[nodep] (fc) at (1.10,2.55) {C};
\draw[arr] (fa) -- (fb);
\draw[arr] (fb) -- (fc);
\node[sub, anchor=west] at (-1.55,1.92) {propagate dependencies};

\node[card] at (3.85,2.75) {};
\node[title, anchor=west] at (2.30,3.58) {Intersection query};
\node[sub, anchor=west] at (2.30,3.30) {Scattered set};
\node[nodeg] (ic1) at (2.70,2.95) {$C_1$};
\node[nodeg] (ic2) at (3.85,2.35) {$C_2$};
\node[nodeg] (ic3) at (5.00,2.95) {$C_3$};
\node[noder] (ix) at (3.85,2.95) {$x$};
\draw[thinarr] (ic1) -- (ix);
\draw[thinarr] (ic2) -- (ix);
\draw[thinarr] (ic3) -- (ix);
\node[sub, anchor=west] at (2.30,1.92) {satisfy all constraints};

\node[card] at (7.70,2.75) {};
\node[title, anchor=west] at (6.15,3.58) {Comparative reasoning};
\node[sub, anchor=west] at (6.15,3.30) {Parallel matrix};
\draw[draw=black!35, line width=.45pt] (6.60,2.20) rectangle (8.80,3.05);
\draw[draw=black!35, line width=.45pt] (7.70,2.20) -- (7.70,3.05);
\draw[draw=black!35, line width=.45pt] (6.60,2.62) -- (8.80,2.62);
\node[sub] at (7.15,2.84) {$v_{11}$};
\node[sub] at (8.25,2.84) {$v_{12}$};
\node[sub] at (7.15,2.40) {$v_{21}$};
\node[sub] at (8.25,2.40) {$v_{22}$};
\node[sub, anchor=west] at (6.15,1.92) {align entities by attributes};

\node[card] at (11.55,2.75) {};
\node[title, anchor=west] at (10.00,3.58) {Temporal reconstruction};
\node[sub, anchor=west] at (10.00,3.30) {Temporal scatter};

\node[nodeo] (t3) at (10.20,2.78) {$T_3$};
\node[nodeo] (t1) at (11.10,2.78) {$T_1$};
\node[nodeo] (t4) at (12.00,2.78) {$T_4$};
\node[nodeo] (t2) at (12.90,2.78) {$T_2$};

\draw[thinarr]
  (t1.south) to[out=-45,in=-135,looseness=0.5] (t2.south);

\draw[thinarr]
  (t2.south) to[out=-115,in=-35,looseness=0.7] (t3.south);

\draw[thinarr]
  (t3.north) to[out=45,in=145,looseness=0.4] (t4.north);

\node[sub, anchor=west] at (10.00,1.92) {doc order $\neq$ story order};

\node[bottomcard] at (0,0) {};
\node[title, anchor=west] at (-1.55,.94) {Causal attribution};
\node[sub, anchor=west] at (-1.55,.66) {Fan-in};
\node[nodeg] (ca1) at (-1.10,.29) {$F_1$};
\node[nodeg] (ca2) at (-1.10,-.27) {$F_2$};
\node[nodeg] (ca3) at (-1.10,-.83) {$F_3$};
\node[noder] (ce) at (.70,-.27) {$E$};
\draw[arr] (ca1) -- (ce);
\draw[arr] (ca2) -- (ce);
\draw[arr] (ca3) -- (ce);
\node[sub, anchor=west] at (-1.55,-1.16) {aggregate contributors};

\node[bottomcard] at (3.85,0) {};
\node[title, anchor=west] at (2.30,.94) {Abductive inference};
\node[sub, anchor=west] at (2.30,.66) {Reverse fan-in};
\node[nodeo] (ao1) at (2.70,.29) {$O_1$};
\node[nodeo] (ao2) at (2.70,-.27) {$O_2$};
\node[nodeo] (ao3) at (2.70,-.83) {$O_3$};
\node[noder] (ah) at (4.65,-.27) {$H$};

\draw[thinarr] (ah) -- (ao1);
\draw[thinarr] (ah) -- (ao2);
\draw[thinarr] (ah) -- (ao3);

\node[sub, anchor=west] at (2.30,-1.16) {infer latent explanation};

\node[bottomcard, minimum width=7.42cm] at (9.66,0) {};
\node[title, anchor=west] at (6.35,.94) {Counterfactual reasoning};
\node[sub, anchor=west] at (6.35,.66) {Branching dependency};
\node[nodep] (cx) at (6.85,-.09) {$X$};
\node[nodep] (cb) at (8.05,-.09) {$B$};
\node[noder] (cy) at (9.25,-.09) {$Y$};
\node[nodeo] (cnb) at (8.05,-.81) {$\neg B$};
\node[nodeo] (cny) at (9.25,-.81) {$\neg Y$};
\draw[arr] (cx) -- (cb);
\draw[arr] (cb) -- (cy);
\draw[thinarr] (cx) -- (cnb);
\draw[thinarr] (cnb) -- (cny);
\node[sub, anchor=west] at (10.25,-.07) {factual path};
\node[sub, anchor=west] at (10.25,-.79) {alternative path};
\node[sub, anchor=west] at (6.35,-1.16) {evaluate what would change};

\end{tikzpicture}%
}

\caption{\textbf{Seven source-induced evidence geometries.} Each schematic node is a grounded source span and each edge is a source-internal relation discovered from the document rather than inserted by the benchmark. The geometry label names the primary dependency required for rubric credit: chains propagate constraints, intersections satisfy scattered conditions, comparisons align attributes, temporal items recover event order, causal items aggregate contributors, abductive items infer latent explanations, and counterfactual items preserve an actual path plus an alternative branch.}
\label{fig:evidence-geometries}
\end{figure}

The seven evidence geometries specify the answer-critical dependency, not a surface template: chains propagate constraints; intersections satisfy scattered conditions; comparisons align a shared dimension; temporal items recover event order; causal items aggregate contributors to an outcome; abductive items infer a latent explanation from observations; and counterfactual items preserve an actual path plus an alternative branch. Labels are primary diagnostic labels at the item-and-rubric level, so ambiguous secondary relations are recorded but do not change the scored relation. Figure~\ref{fig:evidence-geometries} gives the taxonomy; Appendix~\ref{app:source-geometry-artifacts} reports label reliability.


\subsection{Source-first Trail Induction}

\systemName{} is constructed source-first rather than question-first. We do not write a desired multi-hop question and then search for or synthesize supporting passages. Candidate generation begins by asking whether the raw source already contains a supported evidence trail with a clear dependency structure. This order is central to the benchmark: clue placement, relation type, and distractors are properties of the document before they become properties of an evaluation item.

Mechanically, long documents are segmented into overlapping windows, converted into local evidence records, and linked into a typed source-internal graph. Rules and LLM-assisted extraction propose entities, events, attributes, temporal markers, quotations, and local relations; an edge is retained only when the endpoint facts are source-supported and the typed relation is explicit or reviewably entailed by the surrounding windows.


Geometry-specific graph queries then mine candidate trails. Only after a supported trail exists do we write the question, answer, key facts, and rubric. Questions expose enough information to be well-posed without revealing the hidden evidence relation, and rubrics score answer-critical relations rather than surface similarity. Appendix~\ref{app:source-first-induction} gives the artifact schema, induction pseudocode, and the rule/LLM/review boundary.

\subsection{Validity Gates and Release Promotion}

Natural evidence trails are easy to over-generate and hard to validate. A candidate may look multi-hop while one clue already determines the answer; a publicly familiar source may make the question answerable without the document; a proposed causal bridge may be plausible but unsupported; or a nominal geometry may collapse into a simpler local extraction task. The release-promotion checks therefore implement a threat model for benchmark validity rather than a generic quality score.

We organize promotion into four families of validity checks, with the individual checks numbered D1--D9 in the audit artifact and Appendix~\ref{app:strict-promotion-gates}. \emph{Source grounding} requires every answer-critical fact and retained clue to be supported by and relevant to the source. \emph{Multi-hop necessity and shortcut resistance} uses subset, locality, dispersion, and leave-one-out checks to reject local bypasses, shortcut edges, and single-clue-sufficient items. \emph{Contamination resistance} probes whether the question can be answered from public familiarity or memorized summaries without the source. \emph{Answerability and rubric fidelity} verifies that strong readers can answer when the hidden evidence windows are supplied and that the rubric scores the answer core rather than decorative details.

The checks are binding rather than ceremonial. No-document probing rejects 38\% of the validation-tracked pool; leave-one-out and single-clue audits reject items where one clue alone determines the unique answer. After all filters and audits, 3,506 tracked candidates yield 481 locked tasks, a 13.7\% survival rate. Appendix~\ref{app:strict-promotion-gates} gives the D1--D9 checks and grouped pass rules.

\begin{figure}[!htbp]
  \centering
  \includegraphics[width=\textwidth]{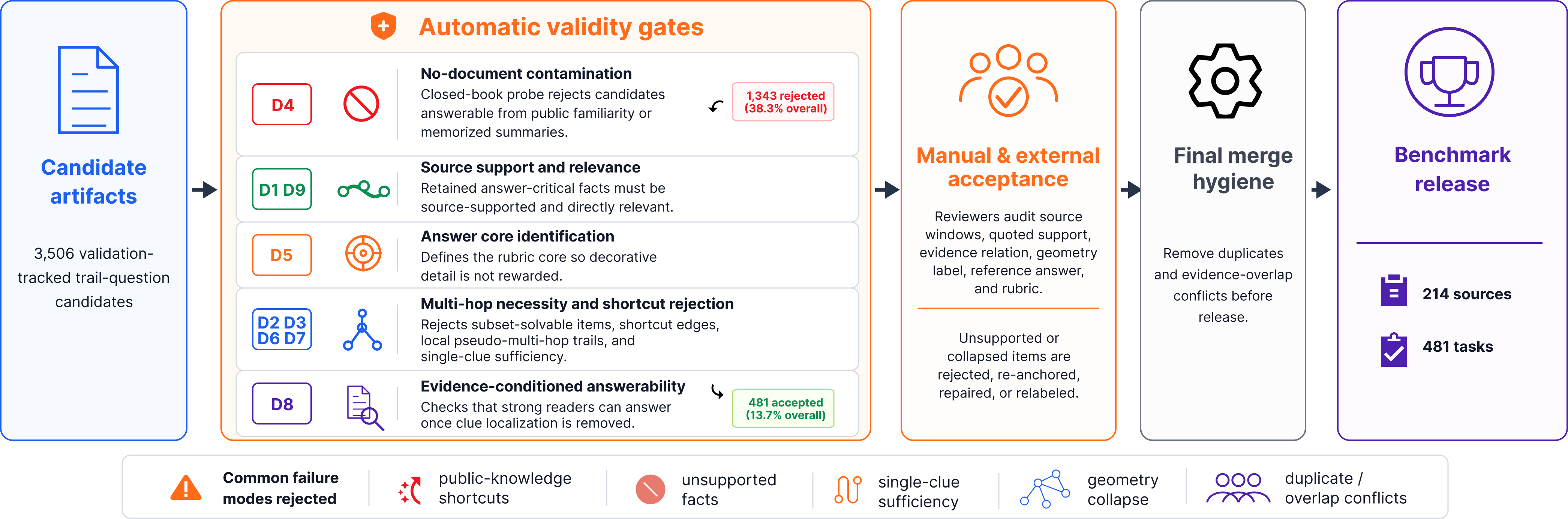}
  \caption{\textbf{Strict promotion and validation funnel.}
  Candidate artifacts enter the locked release only after no-document
  contamination probes, source-support and relevance checks, answer-core review,
  multi-hop necessity and shortcut rejection, evidence-conditioned
  answerability, manual/external acceptance, and final merge hygiene. The
  funnel separates construction-time validation artifacts from the evidence-
  withheld evaluation setting.}
  \label{fig:validation-funnel}
\end{figure}

Automatic checks are followed by manual and external acceptance. Reviewers inspect the source windows, quoted support, evidence relation, geometry label, reference answer, and rubric; failures trigger rejection, evidence re-anchoring, rubric repair, or relabeling rather than silent score discounting. A blinded relabeling check confirms that original geometry labels remain reproducible at \(\kappa=0.88\)--0.89 (Appendix~\ref{app:source-geometry-artifacts}). Final merge hygiene removes duplicates and evidence-overlap conflicts before an item enters the locked release. Figure~\ref{fig:validation-funnel} summarizes the main promotion funnel.

\subsection{Full-Document Evaluation and Rubric Scoring}
\label{sec:evaluation-protocol}

Evaluation follows the evidence-withheld boundary summarized in Appendix Table~\ref{tab:evaluation-boundary}. The evaluated model receives only the public question and the complete source document; all construction and scoring artifacts remain hidden from the model.

Open-form responses are graded against criterion-level rubrics by three non-contestant judges---Claude-Sonnet-4.6, Qwen3.5-Plus, and Gemini-2.5-Flash---and the task score is their mean credit. The rubrics reward
answer-critical facts and relations rather than exact-match or span overlap. Production judges see the question, rubric, and response, but not the source;
rubric criteria are source-verified during item construction and audited through validation and external review. Appendices~\ref{app:strict-promotion-gates}
and~\ref{app:evaluation-boundary} report role separation, judge agreement, evidence-visible checks, and expert validation. The reported score panel uses the frozen release score matrix, and the public release provides the locked task set, validation metadata, and score matrices.

\FloatBarrier

\section{Results}
\label{sec:results}

The leaderboard establishes the overall level of performance, while the diagnostic views reveal how that performance is structured. We analyze two complementary axes: document scale, which expands the source over which relevant evidence must be located and maintained, and evidence geometry, which determines how the selected evidence must be composed into a complete answer.

\subsection{Frontier Performance Reveals Substantial Headroom}

\textbf{\systemName{} remains meaningfully unsaturated at the frontier.} Table~\ref{tab:leaderboard} summarizes performance across the 18 evaluated
systems. The strongest system reaches 75.3\% mean rubric credit, preserving 24.7 points of headroom to full credit.

Criterion-level scoring makes this headroom informative. Each response earns credit for recovered answer-critical facts and relations, while full credit
requires the complete evidence structure encoded by the rubric. The resulting score captures analytical completeness: it distinguishes answers that recover
relevant content from answers that successfully assemble the requested comparison, chronology, explanation, or branch. The leaderboard therefore
summarizes how fully each system converts a long source into a warranted answer.

\begin{table}[t]
\centering
\scriptsize
\setlength{\tabcolsep}{2.4pt}
\renewcommand{\arraystretch}{0.96}
\caption{\textbf{Evidence-withheld full-document score panel.} Scores are frozen three-judge rubric means from the benchmark-release score matrix; bold rows overlap the leader's 95\% task-bootstrap interval.}
\label{tab:leaderboard}
\begin{tabular}{@{}r l r c@{\hspace{1.1em}}r l r c@{}}
\toprule
Rank & Model & Score & CI & Rank & Model & Score & CI \\
\midrule
1 & \textbf{Gemini-3.1-Pro} & \textbf{75.3} & [72.7, 77.7] &
10 & Gemini-2.5-Pro & 61.4 & [58.5, 64.2] \\
2 & \textbf{Claude Opus 4.8} & \textbf{72.6} & [69.7, 75.3] &
11 & DeepSeek-V4 & 60.4 & [57.6, 63.2] \\
3 & \textbf{GPT-5.5} & \textbf{71.0} & [68.3, 73.7] &
12 & GPT-4.1 & 60.4 & [57.6, 63.2] \\
4 & \textbf{GLM-5.2} & \textbf{70.9} & [68.2, 73.6] &
13 & GPT-5.1 & 42.1 & [38.8, 45.2] \\
5 & \textbf{Qwen3.7-Max} & \textbf{70.1} & [67.4, 72.8] &
14 & Qwen3-Max & 35.2 & [31.7, 38.6] \\
6 & Claude Opus 4.6 & 69.2 & [66.5, 71.9] &
15 & Kimi-K2.6 & 35.1 & [31.4, 38.6] \\
7 & Qwen3.6+ & 66.7 & [63.8, 69.5] &
16 & DeepSeek-V3.2 & 33.0 & [30.1, 35.9] \\
8 & Qwen3.7-Plus & 65.7 & [63.0, 68.4] &
17 & Doubao-Seed-2.1 & 30.7 & [26.9, 34.3] \\
9 & GPT-5.4 & 62.2 & [59.4, 65.1] &
18 & MiniMax-M2.7 & 28.9 & [25.5, 32.4] \\
\bottomrule
\end{tabular}
\end{table}

\subsection{Performance Changes Systematically with Document Scale}

\textbf{Full-document reasoning becomes increasingly demanding as the source grows.} Figure~\ref{fig:diagnostic-landscape}A reports performance across the powered L0--L6 context tiers. Most systems achieve their strongest results on the
shortest tier and lower scores on the longest powered tiers. The leading systems nevertheless retain substantial performance through L6, producing a
graded decline rather than an abrupt context cliff. The shape of the curves adds an important second insight. Several systems show local rebounds across adjacent tiers, indicating that source length and
evidence organization jointly shape performance. Longer documents broaden the space of plausible passages and extend the span over which the question, partial findings, and cross-passage relations must be maintained. At the same time, coherent evidence trails can remain recoverable at long scales when their clues reinforce a clear relation. \systemName{} therefore captures not only context reach, but the ability to sustain goal-directed evidence integration as the source grows. The three L7 items are reserved as \(>\)1M-token stress probes and are excluded from the powered tier trend.

\subsection{Evidence Geometry Separates Distinct Reasoning Demands}

\begin{figure}[h]
\centering
\includegraphics[width=0.98\linewidth]{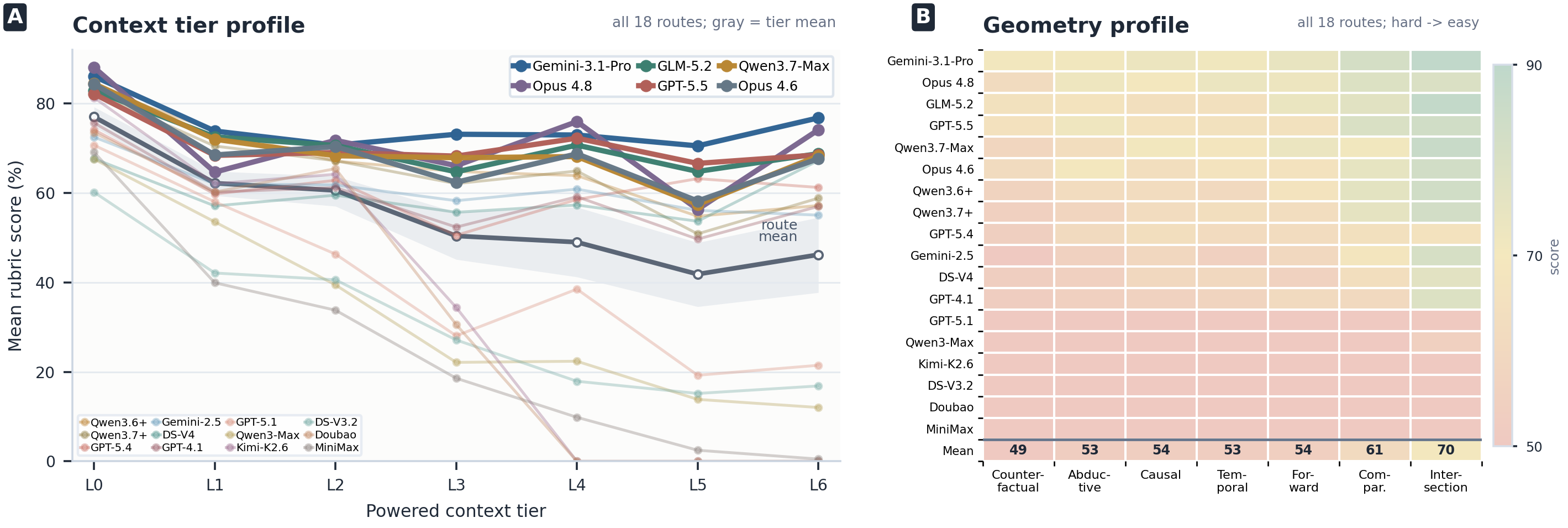}
\caption{\textbf{Length and geometry provide complementary diagnostics.} Performance by context tier and evidence geometry. Panel A reports mean rubric credit across the powered L0--L6 context tiers.
Panel B reports mean credit by primary evidence geometry. L7 is excluded from tier comparisons because it contains only three stress-test items.}
\label{fig:diagnostic-landscape}
\label{fig:context-length}
\label{fig:geometry-profile}
\end{figure}

\textbf{The relation encoded by an evidence trail is a major axis of performance.} Figure~\ref{fig:diagnostic-landscape}B reveals a pronounced geometry profile.
Averaged across the evaluated systems over L0--L6, intersection items reach 70\% mean rubric credit and comparative items reach 61\%, while counterfactual items average 49\%. Forward, causal, temporal, and abductive items form an intermediate band at approximately 53--54\%. The 21-point spread between intersection and counterfactual makes evidence geometry a central diagnostic dimension of \systemName{}.

The ordering reflects the structural operation required by each geometry. Intersection items accumulate multiple compatible constraints toward a single
entity, event, or conclusion. Comparative items align two cases along a shared dimension, providing a natural scaffold for organizing evidence. Forward chains require intermediate constraints to be carried across successive steps, while temporal items require event order to be reconstructed when it differs from document order. Causal and abductive items require the answer to
preserve explanatory direction between observations, contributors, and outcomes. Counterfactual reasoning places the strongest branch-management demand on the model. A complete answer must maintain the factual and alternative states
simultaneously, identify the point at which they diverge, and propagate only the consequences licensed by the alternative branch. This combination of
state separation and dependency tracking helps explain why counterfactual items remain the most challenging geometry in the aggregate profile.

These geometry profiles turn the overall score into a capability signature. Systems with similar aggregate performance can exhibit different strengths in
constraint convergence, cross-case alignment, temporal reconstruction, explanatory attribution, and branch-sensitive reasoning. Document scale defines how far the model must search and retain; evidence geometry defines how the recovered information must be composed.

\section{Discussion}
\label{sec:discussion}

\textbf{The central object is a question-conditioned evidence state.}
The results support a shift from viewing long-context capability as token access to viewing it as the construction of a question-conditioned evidence
state. A successful system must transform a long source into a compact state that preserves answer-critical facts, entity bindings, and relations. Document
scale expands the material that must be filtered, while evidence geometry determines the structure that must survive this filtering. For an intersection
item, this state is a set of converging constraints; for a comparison, a pair of cases aligned along a shared dimension; for temporal reconstruction, an
event order distinct from discourse order; and for counterfactual reasoning, separated
factual and alternative branches. This perspective connects the length and geometry results: effective long-context reasoning depends on selective retention with relational fidelity.

\textbf{Different long-context architectures instantiate different forms of
effective memory.}
Publicly documented systems already span interleaved local/global attention, latent KV compression, hybrid compressed sparse attention, and compressive
memory
\cite{gemmateam2025gemma3,
munkhdalai2024infiniattention,deepseekai2026deepseekv4}.
These mechanisms allocate long-range computation differently: global access supports direct interaction among distant tokens; local/global hybrids
concentrate that interaction in selected layers; compressed states retain compact representations of earlier context. \systemName{} provides a natural
behavioral probe for these trade-offs. Intersection and comparative items test whether multiple anchors remain jointly addressable, temporal and causal items
test preservation of order and direction, and counterfactual items test whether factual and alternative states remain distinct. Controlled studies with open-weight models could pair architecture and cache or memory ablations with geometry-specific scores, connecting systems design to reasoning behavior.

\textbf{Internal and external memory should be studied jointly.}
Attention and compression determine what survives within a model call, while a harness determines what is revisited and restored across calls. Their
interaction can be geometry-specific: a summary may preserve salient facts yet blur event order, causal direction, or branch identity, whereas structured
evidence memory can keep these dependencies explicit. This makes model--harness co-adaptation a central future direction. Comparing direct prompting, shared harnesses, and harness-aware post-training under matched
inference budgets---and testing transfer across harnesses---would separate general gains in long-context reasoning from adaptation to a particular
orchestration policy.

\textbf{The key objective is relation-preserving compression.}
A useful memory need not reproduce the document; it must preserve the evidence state sufficient for the question. Training can therefore supervise
intermediate structures such as entity bindings, comparison axes, temporal order, causal dependencies, and factual-versus-alternative branches, in
addition to final answers. \systemName{} offers a natural testbed for measuring how these structures survive as context length, memory budget, and compression increase. The desired behavior is graceful forgetting: systems should discard redundant surface detail before answer-critical dependencies. By pairing
natural evidence trails with explicit relational structure, \systemName{} provides a common substrate for connecting architecture, memory policy, post-training, and orchestration to this objective.

\section{Limitations}
\label{sec:limitations}

The main limitations of \systemName{} concern coverage and experimental scope. The current release focuses on public technical incident reports and literary
narratives, with English sources forming the majority and Chinese literature providing an initial non-English subset. Broader conclusions across legal,
medical, scientific, code, and enterprise documents will require additional domain- and language-specific collections. The seven evidence geometries assign
one primary answer-critical relation to each task, whereas real analytical questions may combine several relations within a single evidence structure. Extending the benchmark to explicitly multi-geometry tasks is therefore an important direction. Finally, the present experiments focus on direct full-document inference; retrieval-augmented, iterative, and harness-aware
systems form a complementary setting for future evaluation.

\clearpage
\bibliography{references}

\clearpage
\appendix
\section{Additional Construction and Validation Details}
\label{app:additional-construction}

This appendix begins with related work and benchmark positioning, then gives the operational details behind Section~\ref{sec:benchmark-construction}. Figure~\ref{fig:evidence-geometries} in the main text gives the source-internal evidence-geometry taxonomy, and Figure~\ref{fig:validation-protocol} summarizes the release-promotion loop. We then spell out the artifact record, validation gates, manual-review procedure, role separation, and evaluation boundary used to audit the final benchmark.

\subsection{Related Work and Benchmark Positioning}
\label{app:benchmark-positioning}

Table~\ref{tab:related-benchmarks} is a boundary map, not a ranking of benchmark quality. Its columns ask whether a feature is a primary evaluation focus under the following operational definitions. ``Raw-source evaluation'' means that the evaluated unit is a naturally occurring long document rather than a benchmark-assembled passage set. ``Same-source evidence'' means the answer-critical evidence is distributed within one source. ``Evidence hidden at test'' means the model is not given supporting spans or clue windows at evaluation time. ``Shortcut / necessity audit'' means the benchmark checks whether the answer can be recovered from a single local clue or shortcut edge. ``Relation taxonomy'' means the dataset labels the primary evidence relation being tested, not merely the task domain.

\paragraph{Long-context evaluation.}
Long-context evaluation has moved from broad coverage toward sharper tests of realistic reasoning. Earlier suites such as LongBench, L-Eval, InfiniteBench, LooGLE, HELMET, and LOONG established diverse long-input tasks across QA, summarization, few-shot learning, code, and multitask reasoning~\NoHyper\citep{bai2024longbench,an2024leval,zhang2024infinitebench,li2024loogle,yen2024helmet,wang2024loong}\endNoHyper. Newer suites emphasize adjacent axes: LongBench v2 and LongBench Pro stress harder realistic long-context tasks, 100-LongBench audits whether de facto long-context tests isolate long-context ability, and BRIDGE and DocScope add step-, page-, region-, and fact-level evidence supervision for long-document reasoning~\NoHyper\citep{bai2025longbenchv2,chen2026longbenchpro,yang2025hundredlongbench,xiang2026bridge,feng2026docscope}\endNoHyper. Controlled probes such as Lost-in-the-Middle, RULER, BABILong, NeedleBench, and NoLiMa further isolate position sensitivity, needle retrieval, synthetic fact tracing, aggregation, and lexical-overlap effects~\NoHyper\citep{liu2024lost,hsieh2024ruler,kuratov2024babilong,li2024needlebench,modarressi2025nolima}\endNoHyper. These benchmarks are crucial for measuring access and robustness. In many of these settings, however, key parts of the evidence environment are supplied, inserted, supervised, or otherwise controlled by the benchmark rather than discovered from a raw source.

\paragraph{Multi-hop and long-document QA.}
Multi-hop QA datasets such as HotpotQA, 2WikiMultiHopQA, and MuSiQue introduced bridge and comparison questions, supporting facts, reasoning paths, and filters against shortcut-solvable composition~\NoHyper\citep{yang2018hotpotqa,ho2020constructing,trivedi2022musique}\endNoHyper. Long-document QA benchmarks such as NarrativeQA, Qasper, QuALITY, NovelQA, MIR-Bench, SCALAR, and AcademicEval move toward longer books, papers, scripts, and domain-grounded tasks~\NoHyper\citep{kocisky2018narrativeqa,dasigi2021qasper,pang2022quality,wang2025novelqa,yan2025mirbench,wang2026scalar,zhang2025academiceval}\endNoHyper. \systemName{} inherits the need for supporting evidence and shortcut resistance, but changes the origin of the reasoning path: candidate paths are mined from the source's own narrative, temporal, causal, and entity-relation structure before a question is written.

\paragraph{Retrieval and citation-grounded generation.}
Retrieval-augmented and citation-grounded systems make provenance explicit: the model retrieves, cites, or verifies passages as part of the task interface~\NoHyper\citep{lewis2020rag,gao2023alce,liu2023verifiability}\endNoHyper. \systemName{} evaluates a different boundary. It withholds all evidence metadata and asks whether a model can infer which naturally occurring passages are evidential in the first place. This makes the task a test of source-internal evidence selection and relation preservation, not only of using retrieved support.

\paragraph{Frontier benchmark design.}
Recent frontier evaluations increasingly emphasize unsaturated tasks, expert or workflow-grounded construction, and verifiable or rubric-based scoring. Humanity's Last Exam uses expert-vetted closed-ended questions at the frontier of academic knowledge~\NoHyper\citep{phan2025hle}\endNoHyper; BrowseComp focuses on hard-to-find web facts with short verifiable answers~\NoHyper\citep{wei2025browsecomp}\endNoHyper; DeepResearch Bench evaluates long-form research reports and citation quality~\NoHyper\citep{du2025deepresearch}\endNoHyper; Terminal-Bench 2.0 and Odysseys evaluate realistic long-horizon agent workflows with execution tests or graded rubrics~\NoHyper\citep{merrill2026terminalbench,jang2026odysseys}\endNoHyper. \systemName{} follows the same validity-first direction, but targets a different object of measurement: evidence relations that are internal to a single long source and hidden from the evaluated model.

\paragraph{Positioning.}
\systemName{} is complementary to existing long-context, multi-hop, narrative-QA, and retrieval-grounded benchmarks. Prior suites cover important parts of the design space; \systemName{} isolates a particular conjunction as its primary evaluation unit: a single raw long-form source whose clues, distractors, and dependencies arise internally, with evidence hidden from the evaluated model. Success requires preserving the evidence relation that warrants the answer, not only locating a fact or producing a plausible response.

\definecolor{WTYes}{HTML}{2457A6}
\definecolor{WTPart}{HTML}{8A6A18}
\definecolor{WTNo}{HTML}{7A8594}

\providecommand{\wtyes}{}
\providecommand{\wtpart}{}
\providecommand{\wtno}{}
\renewcommand{\wtyes}{\textcolor{WTYes}{\textsf{\scriptsize P}}}
\renewcommand{\wtpart}{\textcolor{WTPart}{\textsf{\scriptsize part.}}}
\renewcommand{\wtno}{\textcolor{WTNo}{\textsf{\scriptsize --}}}

\newcolumntype{C}[1]{>{\centering\arraybackslash}m{#1}}

\begin{table*}[t]
\centering
\scriptsize
\setlength{\tabcolsep}{1.0pt}
\renewcommand{\arraystretch}{0.98}

\resizebox{\textwidth}{!}{%
\begin{tabular}{@{}p{3.55cm}p{3.35cm}ccccc@{}}
\toprule
\textbf{Benchmark / resource} &
\multicolumn{1}{c}{\textbf{Unit}} &
\makecell{\textbf{Raw-source}\\\textbf{eval.}} &
\makecell{\textbf{Same-source}\\\textbf{evidence}} &
\makecell{\textbf{Evidence hidden}\\\textbf{at test}} &
\makecell{\textbf{Shortcut /}\\\textbf{necessity audit}} &
\makecell{\textbf{Relation}\\\textbf{taxonomy}} \\
\midrule

\multicolumn{7}{@{}c@{}}{\emph{Broad and application-centric long-context suites}} \\
\midrule
LongBench~\citep{bai2024longbench}
& Long-context task suite
& \wtpart & \wtpart & \wtpart & \wtpart & \wtno \\

L-Eval~\citep{an2024leval}
& Long-document evaluation
& \wtyes & \wtpart & \wtpart & \wtpart & \wtno \\

$\infty$Bench~\citep{zhang2024infinitebench}
& 100K+ context suite
& \wtpart & \wtpart & \wtpart & \wtpart & \wtno \\

LooGLE~\citep{li2024loogle}
& Long-dependency QA
& \wtyes & \wtpart & \wtyes & \wtpart & \wtno \\

HELMET~\citep{yen2024helmet}
& Application-centric suite
& \wtpart & \wtpart & \wtpart & \wtpart & \wtno \\

LongBench v2~\citep{bai2025longbenchv2}
& Realistic long-context multitask
& \wtyes & \wtpart & \wtpart & \wtpart & \wtno \\

100-LongBench~\citep{yang2025hundredlongbench}
& Length-controlled evaluation
& \wtpart & \wtpart & \wtpart & \wtpart & \wtno \\

\midrule
\multicolumn{7}{@{}c@{}}{\emph{Synthetic, haystack, and controlled long-context probes}} \\
\midrule
RULER~\citep{hsieh2024ruler}
& Controlled context probe
& \wtno & \wtno & \wtyes & \wtpart & \wtno \\

BABILong~\citep{kuratov2024babilong}
& Reasoning-in-a-haystack
& \wtno & \wtno & \wtyes & \wtpart & \wtno \\

NeedleBench~\citep{li2024needlebench}
& Needle tracing / retrieval
& \wtno & \wtno & \wtyes & \wtpart & \wtno \\

NoLiMa~\citep{modarressi2025nolima}
& Lexical-gap needle retrieval
& \wtno & \wtno & \wtyes & \wtno & \wtno \\

\midrule
\multicolumn{7}{@{}c@{}}{\emph{Compositional multi-hop QA benchmarks}} \\
\midrule
HotpotQA~\citep{yang2018hotpotqa}
& Multi-hop QA
& \wtno & \wtno & \wtpart & \wtpart & \wtpart \\

2WikiMultiHopQA~\citep{ho2020constructing}
& Multi-hop QA
& \wtno & \wtno & \wtpart & \wtyes & \wtpart \\

MuSiQue~\citep{trivedi2022musique}
& Composed multi-hop QA
& \wtno & \wtno & \wtpart & \wtyes & \wtpart \\

\midrule
\multicolumn{7}{@{}c@{}}{\emph{Long-document, narrative, and domain-QA benchmarks}} \\
\midrule
NarrativeQA~\citep{kocisky2018narrativeqa}
& Book / script QA
& \wtyes & \wtpart & \wtyes & \wtpart & \wtno \\

QASPER~\citep{dasigi2021qasper}
& Scientific-paper QA
& \wtyes & \wtpart & \wtpart & \wtpart & \wtno \\

QuALITY~\citep{pang2022quality}
& Long-passage QA
& \wtyes & \wtpart & \wtyes & \wtpart & \wtno \\

NovelQA~\citep{wang2025novelqa}
& Novel QA
& \wtyes & \wtpart & \wtyes & \wtpart & \wtno \\

\midrule
\multicolumn{7}{@{}c@{}}{\emph{Source-internal evidence benchmark}} \\
\midrule
\systemName{}
& Source-internal evidence item
& \wtyes & \wtyes & \wtyes & \wtyes & \wtyes \\

\bottomrule
\end{tabular}}

\caption{\textbf{Evaluation-boundary map for representative long-context, multi-hop, and document-QA benchmarks.}
The table compares each benchmark only with respect to the boundary targeted by \systemName{}, not overall benchmark quality or the value of other task designs.
\wtyes{} marks a primary evaluation focus, \wtpart{} marks partial, task-specific, or indirect coverage, and \wtno{} means the axis is not a primary focus under the operational definitions in Appendix~\ref{app:benchmark-positioning}.
\systemName{} combines these axes in one setting: naturally distributed source evidence, evidence-hidden full-document evaluation, necessity validation, and a relation-level taxonomy of reasoning demands.}
\label{tab:related-benchmarks}
\end{table*}

\subsection{Source, Geometry, and Candidate Artifacts}
\label{app:source-geometry-artifacts}

\paragraph{Source selection and tiering.}
The benchmark release uses 214 long-form source files from three intentionally different source families: public English technical incident reports, English literary narratives, and Chinese literary sources. These families stress different evidence regimes. Technical reports make causal, procedural, and corrective-action relations explicit but require preserving operational dependencies. Literary sources make many relations implicit through delayed revelation, perspective, discourse order, and motive. Chinese sources provide an initial Chinese literature slice and diversify the source regime beyond English texts. Table~\ref{tab:context-tier-coverage} reports how the 481 locked items are distributed across full-document length tiers.

\begin{table}[!htbp]
\centering
\footnotesize
\setlength{\tabcolsep}{4.2pt}
\renewcommand{\arraystretch}{1.12}
\begin{tabularx}{\linewidth}{@{}L{0.34\linewidth} Y@{}}
\toprule
\textbf{Benchmark property} & \textbf{Benchmark release} \\
\midrule
Locked tasks & 481 questions in the benchmark release. \\
Source coverage & 214 long-form source files: English technical reports, English literary narratives, and Chinese literary sources. \\
Geometry counts & Forward chain 73; intersection 71; temporal reconstruction 71; causal attribution 67; comparative 67; counterfactual reasoning 67; abductive inference 65. \\
Context tiers & L0 $\leq$128K: 90; L1 128K--181K: 59; L2 181K--256K: 61; L3 256K--362K: 58; L4 362K--512K: 59; L5 512K--724K: 58; L6 724K--1M: 93. \\
Frontier stress tier & L7 $>$1M: 3 items, reported as a stress probe rather than a powered comparison tier. \\
Sample floor & Fully met: L0/L6 have at least 12 items per geometry, and L1--L5 have at least 8 items per geometry. \\
Release rule & Only items passing release-promotion checks and external acceptance enter the locked release. \\
\bottomrule
\end{tabularx}
\caption{\textbf{What is in the locked benchmark release.} The table summarizes the final dataset after release-promotion checks and external acceptance, including task count, source count, source families, geometry balance, context tiers, and release artifacts. Counts follow the frozen release and audit ledger, not earlier candidate pools.}
\label{tab:dataset-overview-v3}
\end{table}

\begin{table}[H]
\centering
\small
\setlength{\tabcolsep}{4.5pt}
\renewcommand{\arraystretch}{1.12}
\begin{tabularx}{\linewidth}{@{}l l r Y@{}}
\toprule
\textbf{Tier} & \textbf{Length} & \textbf{Items} & \textbf{Reporting role} \\
\midrule
L0 & \(\leq\)128K & 90 & Short full-document tier; at least 12 items per geometry. \\
L1 & 128K--181K & 59 & Lower-mid tier used for degradation analysis; at least 8 items per geometry. \\
L2 & 181K--256K & 61 & Mid tier near common long-context routing boundaries; at least 8 items per geometry. \\
L3 & 256K--362K & 58 & Boundary tier where context handling and relation integration begin to diverge; at least 8 items per geometry. \\
L4 & 362K--512K & 59 & Long tier with hard abductive and branch-heavy cases; at least 8 items per geometry. \\
L5 & 512K--724K & 58 & Long tier; at least 8 items per geometry. \\
L6 & 724K--1M & 93 & Long-context stress tier; at least 12 items per geometry. \\
L7 & \(>\)1M & 3 & Frontier stress probe, not a powered comparison tier. \\
\bottomrule
\end{tabularx}
\caption{\textbf{Full-document length tiers used for analysis.} L0--L6 define the powered benchmark-release length strata and meet the release floor across evidence geometries; L7 contains only three \(>\)1M-token stress-probe items.}
\label{tab:context-tier-coverage}
\end{table}

\FloatBarrier
\clearpage

\begin{figure}[H]
  \centering
  \includegraphics[width=\linewidth]{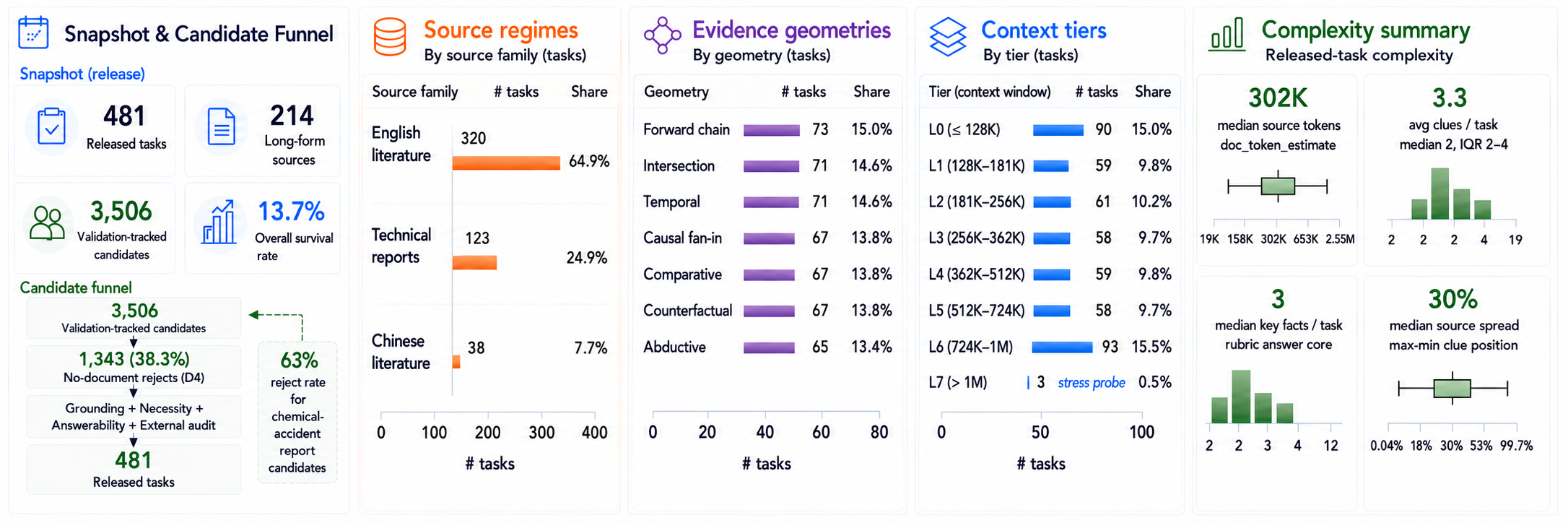}
  \caption{\textbf{\systemName{} release statistics.}
  The dashboard summarizes the locked release: task and source counts,
  validation-tracked candidate funnel, source-family mix, evidence-geometry
  counts, context-tier distribution, and released-task complexity statistics
  computed from the strict481 artifact.}
  \label{fig:release-statistics}
\end{figure}

\paragraph{Primary geometry labels.}
Geometry labels are primary labels, not claims that no other relation appears in the source. Reviewers assign the label according to the dependency that must be preserved to earn the rubric core. For example, a counterfactual branch may contain causal steps, but it remains counterfactual when the answer-critical operation is preserving the alternative non-actual outcome. External acceptance audits reject geometry-collapse cases where the declared label is not the answer-critical relation.

\paragraph{Geometry relabeling reliability.}
We validated this primary-label scheme with a blinded relabeling study over 98 items, stratified 14 per geometry. Two annotators saw only the question, evidence spans, and rubric criteria, not the original label or construction notes. They assigned a primary label and could mark a secondary label when another relation was genuinely present but not answer-critical. The annotators disagreed on 6 of 98 primary labels, giving 93.9\% agreement and Cohen's \(\kappa=0.929\). Agreement against the benchmark's original primary labels was 89.8\% for annotator A (\(\kappa=0.881\)) and 90.8\% for annotator B (\(\kappa=0.893\)); allowing secondary labels raised agreement into the mid-90s. Remaining disagreements mainly reflect adjacent explanatory relations where a secondary relation is present but not the rubric-critical one.

\begin{table}[!htbp]
\centering
\footnotesize
\setlength{\tabcolsep}{4pt}
\renewcommand{\arraystretch}{1.12}
\begin{tabularx}{\textwidth}{@{}L{0.28\textwidth}rrY@{}}
\toprule
\textbf{Reliability check} & \textbf{Agreement} & \textbf{\(\kappa\)} & \textbf{Read} \\
\midrule
Annotator A vs. B primary labels & 93.9\% & 0.929 & Six primary-label disagreements over 98 stratified items. \\
Annotator A vs. original primary & 89.8\% & 0.881 & Original primary labels are broadly reproducible under blinded relabeling. \\
Annotator B vs. original primary & 90.8\% & 0.893 & Primary-or-secondary agreement with the original label reaches the mid-90s. \\
\bottomrule
\end{tabularx}
\caption{\textbf{Blinded evidence-geometry relabeling.} Two annotators assigned primary geometry labels from the question, evidence spans, and rubric, without seeing the original construction label. Agreement is high but not perfect after release filtering. Remaining disagreements mostly reflect primary-versus-secondary choices among adjacent explanatory labels, so geometry slices should be read as diagnostic groupings by the answer-critical relation rather than as mutually exclusive linguistic categories.}
\label{tab:geometry-relabel}
\end{table}

\paragraph{Candidate artifact record.}
Each promoted item is backed by an auditable construction record: source identifier, full-document tier, source-family assignment used for aggregate analysis, intended evidence geometry, selected source windows, grounded evidence spans, induced graph relation, question, reference answer, answer-critical key facts, criterion-level rubric, validation outcomes, and manual-review notes. The evaluated model sees none of these construction artifacts except the raw full source and the question. This separation is the core distinction between \systemName{} and evaluations that provide retrieved passages or benchmark-designed clue layouts.

\paragraph{Artifact schema excerpt.}
The public release uses opaque item identifiers and a clean schema. A full task record contains model-facing fields plus hidden audit fields in the answer-bearing split; internal construction batches, repair notes, and legacy iteration identifiers are not part of the release artifact. The excerpt below illustrates the schema shape without relying on a legacy internal item id or construction log.

\begingroup
\footnotesize
\begin{verbatim}
{
  "question_id": "wildtrace-0001",
  "source_id": "source_technical_001",
  "corpus_file": "corpus/source_technical_001.txt",
  "tier": "L0",
  "source_family": "technical_report",
  "geometry": "abductive",
  "question": "...",
  "evidence_spans": [
    {"span_id": "span_a", "role": "observed outcome", "quote": "..."},
    {"span_id": "span_b", "role": "latent explanation", "quote": "..."}
  ],
  "trail_edges": [
    {"from": "span_b", "to": "span_a", "type": "explains"}
  ],
  "reference_answer": "...",
  "rubric": [
    {"id": "P1", "points": 50, "criterion": "..."},
    {"id": "P2", "points": 50, "criterion": "..."}
  ],
  "validation_summary": "passed release promotion"
}
\end{verbatim}
\endgroup

\subsection{Source-First Induction}
\label{app:source-first-induction}

Item induction proceeds from source structure to question text. First, long documents are segmented into overlapping windows and converted into local evidence records containing entities, events, attributes, temporal markers, relations, quotations, and local summaries. Second, these local records are linked into a document graph only when the surrounding source context supports the relation. Third, geometry-specific graph queries mine candidate paths: conjunctions for intersection, parallel attributes for comparison, text-order/story-order mismatches for temporal reconstruction, fan-in dependencies for causal attribution, reverse fan-in observations for abductive explanation, and factual/alternative branches for counterfactual reasoning. Only after this path exists do we write the question and answer. This prevents the benchmark from becoming a prompt-first search for convenient passages.

\begin{table}[!htbp]
\centering
\footnotesize
\setlength{\tabcolsep}{4pt}
\renewcommand{\arraystretch}{1.13}
\begin{tabularx}{\linewidth}{@{}L{0.12\linewidth}L{0.22\linewidth}Y@{}}
\toprule
\textbf{Stage} & \textbf{Operation} & \textbf{Implementation and validation boundary} \\
\midrule
1 & Segment source & Split raw source \(S\) into overlapping windows with saved offsets, relative positions, source id, and tier metadata. This step is deterministic. \\
2 & Extract local records & For each window, create candidate records for entities, events, attributes, quotations, temporal markers, and local relations. This step is LLM-assisted but every retained record keeps source offsets or quoted support. \\
3 & Propose typed edges & Add candidate edges for entity/coreference continuity, repeated attributes, temporal order, causal language, observation-to-explanation links, factual branches, and comparison dimensions. Rule filters propose high-precision edges; LLM assistance supplies paraphrase and implicit-relation candidates. \\
4 & Verify edge support & Keep an edge only if the endpoint facts are source-supported and the typed relation is explicit or reviewably entailed by the surrounding windows. Unsupported paraphrases, ambiguous pronouns, and thematic rather than causal links are rejected or sent to manual review. \\
5 & Query geometries & Run geometry-specific graph queries to produce candidate trails \(T=(\mathrm{spans},\mathrm{edges},g)\): intersections, comparisons, temporal reconstructions, causal fan-in, abductive explanations, forward chains, and counterfactual branches. \\
6 & Write item artifacts & Write \(q\), \(a\), key facts, and rubric \(R\) from the accepted trail, not before it exists. The public question exposes enough constraints for well-posedness while withholding evidence spans, clue count, graph path, and rubric. \\
7 & Promote or reject & Apply the D1--D9 validation checks, leave-one-out necessity, no-document probing, evidence-conditioned answerability, manual review, and external acceptance. Items answerable from one clue, wrong geometry, or unsupported edges are dropped or repaired before release. \\
\bottomrule
\end{tabularx}
\caption{\textbf{Pipeline overview for source-first trail induction.} The table expands the construction order used by \systemName{}: source \(\rightarrow\) local records \(\rightarrow\) typed graph edges \(\rightarrow\) hidden trail \(\rightarrow\) question/rubric \(\rightarrow\) release gate. LLMs assist extraction and candidate relation proposal, but promotion depends on source-grounded validation, leave-one-out necessity, and human/external review, so the public question is written only after a supported evidence trail exists.}
\label{tab:source-first-pipeline}
\end{table}

\paragraph{Audit-level pseudocode.}
Table~\ref{tab:source-first-pipeline} is a pipeline overview rather than an executable algorithm. The audit procedure used for release promotion is captured more directly by the following pseudocode; implementation scripts may over-generate and repair candidates, but a candidate enters the locked release only through this acceptance boundary.

\begingroup
\footnotesize
\begin{verbatim}
for source S in source_set:
  windows = segment(S, overlapping_windows, keep_offsets=True)
  records = extract_local_records(windows)
  graph = {}
  for candidate_edge in propose_edges(records):
    if source_supports(candidate_edge) and relation_type_is_valid(candidate_edge):
      graph.add(candidate_edge)
  for geometry g in GEOMETRIES:
    for trail T in query_graph(graph, geometry=g):
      item = write_question_answer_rubric(S, T, g)
      if passes_D1_to_D9(item) and passes_manual_external_review(item):
        release.add(lock(item))
\end{verbatim}
\endgroup

\begin{table}[!htbp]
\centering
\footnotesize
\setlength{\tabcolsep}{4pt}
\renewcommand{\arraystretch}{1.12}
\begin{tabularx}{\linewidth}{@{}L{0.23\linewidth}Y@{}}
\toprule
\textbf{Component} & \textbf{Released/auditable specification} \\
\midrule
Segmentation & Nominal 25K-character overlapping induction windows with saved source id, character offsets, relative position, and tier metadata; retained clues additionally store precise offsets and quoted support. \\
Local record extraction & LLM-assisted JSON records over entities, events, attributes, temporal markers, quotations, local relations, and short local summaries; retained records must carry source offsets or quoted evidence. \\
Edge proposal and retention & Rule filters propose high-precision continuity, temporal, comparison, causal, abductive, and branch links; LLM assistance proposes paraphrase or implicit-relation candidates. An edge is retained only when the endpoint facts and relation are supported by the surrounding source windows. \\
Geometry query constraints & Intersection requires a unique entity satisfying all constraints; comparison requires a shared dimension; temporal requires story-time order rather than document order; causal and forward items require a supported dependency direction; abductive items require observations supporting a latent explanation; counterfactual items require both actual and alternative branches. \\
Promotion checks & No-document probing, evidence dispersion, leave-one-out necessity, and evidence-conditioned answerability are applied before release; final acceptance rejects items with unsupported facts, local shortcuts, single-clue sufficiency, or collapsed geometry. \\
Release artifact & Each locked item stores source id, tier, source family, geometry, question, evidence spans, trail edges, reference answer, rubric criteria, validation outcomes, and review notes. Model-facing evaluation hides evidence spans, clue count, graph path, reference answer, rubric, and validation notes. \\
\bottomrule
\end{tabularx}
\caption{\textbf{Construction parameters and audit surface.} The table gives the stable parameters and fields needed to audit construction bias without treating the over-generation code path as a deterministic recipe for reproducing the exact same item set.}
\label{tab:construction-audit-surface}
\end{table}

Graph induction is therefore not a single model judgment. Rules and offsets provide the initial scaffold; LLM-assisted passes propose local records and candidate implicit relations; the source-support, single-clue-sufficiency, and relevance checks (D1, D7, and D9), together with manual review, decide whether the relation is actually grounded in the source. For causal, forward-chain, abductive, and counterfactual items, reviewers also inspect the direction of the dependency: thematic parallels, narrator-stated explanations that collapse abduction, missing alternative outcomes, and single-clue sufficient shortcuts are rejection cases rather than low-confidence edges.

\paragraph{Construction prompt skeletons.}
The release package contains executable loading and scoring utilities; construction-time generation is not part of the public model-evaluation API. To make the annotation boundary auditable, we use short, role-separated prompt skeletons rather than a single end-to-end generator. The local-record prompt asks for source-grounded records only:
\begingroup
\footnotesize
\begin{verbatim}
Given SOURCE_ID, WINDOW_OFFSET, and WINDOW_TEXT, return JSON records:
entities, events, attributes, temporal markers, quotations,
local relations, and a short summary. Every record must include either
a character offset or a quoted passage from WINDOW_TEXT. Do not infer
facts not supported by the window.
\end{verbatim}
\endgroup
The item-writing prompt is invoked only after a supported trail exists:
\begingroup
\footnotesize
\begin{verbatim}
Given SOURCE_ID, GEOMETRY, EVIDENCE_SPANS, and TRAIL_EDGES,
write a public question, reference answer, key facts, and a
criterion-level rubric. The question must not reveal evidence spans,
clue count, graph path, or the reference answer. Each rubric point must
map to an answer-critical fact or relation supported by the spans.
\end{verbatim}
\endgroup

\subsection{Question Writing and Rubric Artifacts}
\label{app:question-rubric-writing}

The question is written to expose enough constraints for the intended reasoning task without leaking the hidden evidence relation. For example, intersection questions may name the constraints that define uniqueness, while abductive and counterfactual questions must avoid naming the missing hypothesis or branch point because recovering that relation is the task. The reference answer and rubric are written from wide context around the evidence spans. Rubric criteria are then reduced to answer-critical facts and relations: facts that are merely vivid, stylistically useful, or locally true but not necessary are removed before scoring. This is why the benchmark gives partial credit for preserved source relations rather than for surface answer similarity.

\subsection{Strict Promotion Gates}
\label{app:strict-promotion-gates}

\paragraph{Gate order.}
Release promotion is dependency ordered. No-document probes run before release to reject contaminated or memorized items. D1--D9 then test whether the answer is grounded, the evidence path is irreducible, the rubric core is precise, and the item remains answerable when evidence localization is provided. The evidence-conditioned ceiling check verifies that strong readers can answer once evidence selection is removed. Dual manual review inspects the artifacts and original source windows before final merge hygiene removes duplicates or evidence-overlap conflicts.

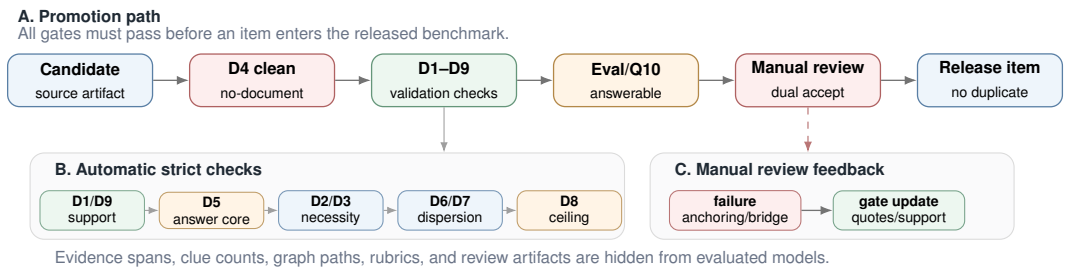
\begin{figure*}[!tbp]
\centering
\definecolor{wtInk}{HTML}{1F2933}
\definecolor{wtMuted}{HTML}{697386}
\definecolor{wtBlue}{HTML}{2F5F8F}
\definecolor{wtBlueLight}{HTML}{EEF5FB}
\definecolor{wtGold}{HTML}{A06A1A}
\definecolor{wtGoldLight}{HTML}{FFF4E6}
\definecolor{wtGreen}{HTML}{3E7C59}
\definecolor{wtGreenLight}{HTML}{EDF7F0}
\definecolor{wtRed}{HTML}{9B3A3A}
\definecolor{wtRedLight}{HTML}{FDEEEE}
\definecolor{wtPanel}{HTML}{FAFAFB}
\resizebox{\linewidth}{!}{%
\begin{tikzpicture}[
  font=\sffamily,
  gate/.style={rounded corners=5pt, line width=.55pt, minimum width=2.15cm,
    minimum height=.78cm, align=center, inner sep=3pt},
  bluegate/.style={gate, draw=wtBlue!80, fill=wtBlueLight},
  greengate/.style={gate, draw=wtGreen!80, fill=wtGreenLight},
  goldgate/.style={gate, draw=wtGold!80, fill=wtGoldLight},
  redgate/.style={gate, draw=wtRed!80, fill=wtRedLight},
  chip/.style={rounded corners=4pt, line width=.45pt, minimum width=1.55cm,
    minimum height=.52cm, align=center, inner sep=2pt, font=\sffamily\tiny},
  panel/.style={draw=black!14, fill=wtPanel, rounded corners=6pt, line width=.45pt},
  arr/.style={-{Latex[length=1.8mm]}, draw=black!55, line width=.55pt},
  softarr/.style={-{Latex[length=1.5mm]}, draw=black!38, line width=.45pt},
  feedback/.style={-{Latex[length=1.6mm]}, draw=wtRed!70, dashed, line width=.50pt}
]

\node[font=\sffamily\bfseries\scriptsize, text=wtInk, anchor=west] at (-1.05,3.98) {A. Promotion path};
\node[font=\sffamily\scriptsize, text=wtMuted, anchor=west] at (-1.05,3.73)
  {All gates must pass before an item enters the released benchmark.};

\node[bluegate]  (cand) at (0,3.05) {\textbf{\scriptsize Candidate}\\[-1pt]\tiny source artifact};
\node[redgate]   (d4)   at (2.70,3.05) {\textbf{\scriptsize D4 clean}\\[-1pt]\tiny no-document};
\node[greengate] (dc)   at (5.40,3.05) {\textbf{\scriptsize D1--D9}\\[-1pt]\tiny validation checks};
\node[goldgate]  (q10)  at (8.10,3.05) {\textbf{\scriptsize Eval/Q10}\\[-1pt]\tiny answerable};
\node[redgate]   (rev)  at (10.80,3.05) {\textbf{\scriptsize Manual review}\\[-1pt]\tiny dual accept};
\node[bluegate]  (keep) at (13.50,3.05) {\textbf{\scriptsize Release item}\\[-1pt]\tiny no duplicate};

\draw[arr] (cand) -- (d4);
\draw[arr] (d4) -- (dc);
\draw[arr] (dc) -- (q10);
\draw[arr] (q10) -- (rev);
\draw[arr] (rev) -- (keep);

\node[panel, minimum width=8.82cm, minimum height=1.28cm, anchor=north west] (auto) at (-.75,1.98) {};
\node[font=\sffamily\bfseries\scriptsize, text=wtInk, anchor=west] at (-.50,1.73) {B. Automatic strict checks};
\node[chip, draw=wtGreen!65, fill=wtGreenLight] (c1) at (.18,1.12) {\textbf{D1/D9}\\support};
\node[chip, draw=wtGold!65, fill=wtGoldLight] (c2) at (1.95,1.12) {\textbf{D5}\\answer core};
\node[chip, draw=wtBlue!65, fill=wtBlueLight] (c3) at (3.72,1.12) {\textbf{D2/D3}\\necessity};
\node[chip, draw=wtBlue!65, fill=wtBlueLight] (c4) at (5.49,1.12) {\textbf{D6/D7}\\dispersion};
\node[chip, draw=wtGold!65, fill=wtGoldLight] (c5) at (7.26,1.12) {\textbf{D8}\\ceiling};
\draw[softarr] (c1) -- (c2);
\draw[softarr] (c2) -- (c3);
\draw[softarr] (c3) -- (c4);
\draw[softarr] (c4) -- (c5);
\coordinate (autoTop) at (5.40,1.98);
\draw[softarr] (dc.south) -- (autoTop);

\node[panel, minimum width=5.40cm, minimum height=1.28cm, anchor=north west] (manual) at (8.45,1.98) {};
\node[font=\sffamily\bfseries\scriptsize, text=wtInk, anchor=west] at (8.70,1.73) {C. Manual review feedback};
\node[chip, minimum width=1.95cm, draw=wtRed!70, fill=wtRedLight] (bad) at (9.72,1.12)
  {\textbf{failure}\\anchoring/bridge};
\node[chip, minimum width=1.95cm, draw=wtGreen!70, fill=wtGreenLight] (hard) at (12.15,1.12)
  {\textbf{gate update}\\quotes/support};
\coordinate (manualTop) at (10.80,1.98);
\draw[feedback] (rev.south) -- (manualTop);
\draw[arr] (bad) -- (hard);

\node[font=\sffamily\scriptsize, text=wtMuted, anchor=west] at (-.50,.40)
  {Evidence spans, clue counts, graph paths, rubrics, and review artifacts are hidden from evaluated models.};

\end{tikzpicture}%
}
\caption{\textbf{Release-promotion protocol and feedback loop.} Candidates must pass contamination probes, D1--D9 strict checks, evidence-conditioned answerability, dual manual review, and final merge hygiene before entering the canonical set. The lower panels summarize review feedback: candidates are rejected or revised when the evidence trail collapses into local extraction, unsupported key facts, contamination, or a mismatched reasoning geometry.}
\label{fig:validation-protocol}
\end{figure*}

\paragraph{D1--D9 grouping.}
The nine dimensions in Table~\ref{tab:validation-dimensions} are not nine independent difficulty knobs. They form four validity groups. D1 and D9 enforce source support and relevance. D5 identifies the answer core so the rubric does not reward decorative detail. D2, D3, D6, and D7 enforce evidence dependence by rejecting shortcut edges, local bypasses, and single-clue sufficiency. D8 tests answerability under evidence-conditioned prompts, while D4/D4-alt handles no-document contamination. This grouping is what Figure~\ref{fig:validation-protocol} expands in its lower-left panel.

\begin{table}[H]
\centering
\footnotesize
\setlength{\tabcolsep}{4pt}
\renewcommand{\arraystretch}{1.14}
\begin{tabularx}{\textwidth}{@{}L{0.05\textwidth} L{0.18\textwidth} L{0.42\textwidth} Y@{}}
\toprule
\textbf{Code} & \textbf{Threat to validity} & \textbf{Operationalization} & \textbf{Pass rule} \\
\midrule
D1 & Unsupported reference answer & Verify each retained rubric fact against the cited source region under wide local context. & All answer-critical facts are source-supported. \\
D2 & Subset-solvable item & Remove each required clue in turn and test whether the intended path still determines the answer. & Removing any required clue breaks the path or leaves the answer underdetermined. \\
D3 & Structural bypass / shortcut edge & Check whether the intended reasoning chain can be collapsed by a direct edge or a single local statement. & No shortcut edge or local bypass remains. \\
D4 & Memorization / contamination & Query strong models without the document and score against the same answer criteria. & No-document probes cannot recover the answer core; alternate probes are used for blocked or ambiguous cases. \\
D5 & Ambiguity and rubric dilution & Use multi-model full-evidence answers to identify the shared answer structure or core facts. Detail-heavy geometries use fact containment; structure-heavy geometries use structural agreement. & Pairwise agreement passes, and non-core padding facts are removed from the answer criteria. \\
D6 & Pseudo-multi-hop from local concentration & Compute evidence spread and minimum gap over clue offsets before expensive validation. & Retained clues must be sufficiently dispersed for the declared geometry; targeted miners may impose stricter spread gates. \\
D7 & Single-clue sufficiency & Re-score each clue window in isolation \emph{against the D5 core facts}, rather than against the original fact list. & No single clue window can recover the answer core or yield the unique full answer by itself. \\
D8 & Unanswerable item or broken rubric & Give strong models all clue windows and ask whether the item is answerable once evidence localization is removed. & Evidence-conditioned readers must be able to recover the answer core from the supplied clue windows. \\
D9 & Tangential or padded rubric facts & Judge whether each grounded fact is necessary or directly relevant to answering the question. & Only relevant facts remain; at least two grounded answer-critical facts survive. \\
\bottomrule
\end{tabularx}
\caption{\textbf{Nine validation dimensions for release promotion.} Each D-check targets a concrete threat to benchmark fidelity: unsupported evidence, shortcut answerability, contamination, rubric dilution, local sufficiency, low ceiling, or mismatched relevance. The protocol is designed to preserve source-internal multi-hop necessity rather than to maximize apparent difficulty.}
\label{tab:validation-dimensions}
\end{table}

\paragraph{D4 no-document probe.}
D4 is an empirical filter for closed-book answerability and memorized shortcuts, not a claim of absolute source novelty. A candidate fails when a strong no-document probe can recover the answer core without the source. This removes candidates whose answer is recoverable from plot familiarity, public summaries, or canonical facts before they can enter the locked release. At the validation-tracked candidate-pool level, D4 rejects 38\% of all candidates, which is why it is treated as a release gate rather than only as a post hoc diagnostic.

\paragraph{Gate-threshold sensitivity.}
The continuous gates are operating checks rather than claims of formal calibration. We use local sensitivity checks to ensure nearby operating points do not qualitatively change the accepted set, and external acceptance review remains binding for borderline cases. D7 is intentionally not summarized as a continuous threshold sweep here: the final release decision uses explicit leave-one-out acceptance, including rejection of any item where one clue alone yields the unique full answer.

\paragraph{Manual review.}
Manual review is a human-led audit of the candidate artifact, assisted where useful by model-generated traces but not delegated to an uninspected model verdict. Reviewers inspect the question, reference answer, key facts, rubric, evidence windows, quoted support, geometry label, and validation failures. A final accepted item must remain grounded, non-local, answerable, and faithful to its declared geometry under this review. The lower-right panel of Figure~\ref{fig:validation-protocol} represents feedback from this review: a failure does not simply lower confidence, it updates the evidence gate by forcing rejection, evidence re-anchoring, rubric repair, or geometry relabeling before a candidate can be reconsidered.

\paragraph{Judge-panel reliability.}
The final scoring panel uses Claude-Sonnet-4.6, Qwen3.5-Plus, and Gemini-2.5-Flash as non-contestant judges. Table~\ref{tab:judge-reliability} reports answer-cell and model-rank agreement, plus a judge-family sensitivity check because Qwen-family systems are also evaluated. Answer-cell Pearson correlations are at least 0.81, pairwise Kendall \(\tau\) over model ranks is at least 0.91, and removing the Qwen judge shifts any model mean by at most 1.8 points.

\begin{table}[H]
\centering
\footnotesize
\setlength{\tabcolsep}{4.5pt}
\renewcommand{\arraystretch}{1.08}
\begin{tabularx}{\linewidth}{@{}L{0.48\linewidth}rY@{}}
\toprule
\textbf{Judge diagnostic} & \textbf{Value} & \textbf{Unit} \\
\midrule
Gemini-2.5-Flash grand mean & 67.4 & score points \\
Claude-Sonnet-4.6 grand mean & 62.3 & score points \\
Qwen3.5-Plus grand mean & 60.9 & score points \\
Sonnet vs. Qwen3.5 Pearson & 0.886 & answer cells \\
Qwen3.5 vs. Gemini-Flash Pearson & 0.817 & answer cells \\
Sonnet vs. Gemini-Flash Pearson & 0.809 & answer cells \\
Sonnet vs. Qwen3.5 Kendall \(\tau\) & 0.987 & model ranks \\
Qwen3.5 vs. Gemini-Flash Kendall \(\tau\) & 0.922 & model ranks \\
Sonnet vs. Gemini-Flash Kendall \(\tau\) & 0.908 & model ranks \\
Qwen judge bias on Qwen-family outputs & -4.06 & pp vs. Sonnet/Gemini mean \\
Qwen judge bias on non-Qwen outputs & -3.88 & pp vs. Sonnet/Gemini mean \\
Qwen-family bias gap & -0.18 & pp \\
Remove-Qwen-judge max model shift & 1.80 & score points \\
All-judge vs. no-Qwen rank Kendall \(\tau\) & 0.974 & model ranks \\
\bottomrule
\end{tabularx}
\caption{\textbf{Final-panel judge reliability on the benchmark release.} The table reports agreement among the three non-contestant judges at two levels: per-answer rubric scores and single-judge model rankings. The Qwen-family bias rows compare Qwen3.5-Plus scores with the mean of the Sonnet and Gemini judges on answer cells scored by all three judges; negative values mean the Qwen judge is stricter, and similar bias on Qwen versus non-Qwen outputs indicates no large observed family-specific preference in this panel.}
\label{tab:judge-reliability}
\end{table}

\paragraph{Expert validation of judge scores.}
We additionally checked whether the rubric panel tracks expert judgment on a stratified 60-response sample. The sample is balanced across the three judge-score bins and the three source families, and covers all seven evidence geometries. Two domain experts scored each response against the same rubric with verbatim key-fact quotes visible; point-level model-assisted pre-scores were used only as drafts, and the experts verified and adjudicated every rubric decision against the evidence. The resulting consensus scores correlate with the three-judge panel at Pearson \(r\approx0.85\), with mean absolute human--judge difference \(\approx0.14\) and mean signed difference \(\approx0\). This supports using the panel mean as a practical proxy for expert rubric scoring, while not replacing larger-scale human rescoring.

\begin{table}[H]
\centering
\footnotesize
\setlength{\tabcolsep}{4.5pt}
\renewcommand{\arraystretch}{1.08}
\begin{tabularx}{\linewidth}{@{}L{0.36\linewidth}Y@{}}
\toprule
\textbf{Validation item} & \textbf{Result} \\
\midrule
Sample size & 60 question--response pairs \\
Score-bin stratification & 20 low, 20 medium, 20 high by three-judge score bin \\
Source-family stratification & 20 technical-report, 20 English-literature, 20 CJK-literature pairs \\
Geometry coverage & All seven evidence geometries, with 7--11 pairs per geometry \\
Expert scoring protocol & Two domain experts verified and adjudicated model-assisted point-level draft labels against the same rubric, with verbatim key-fact quotes visible \\
Human--panel Pearson correlation & \(\approx0.85\) between expert consensus score and three-judge mean \\
Mean absolute difference & \(\approx0.14\) on the normalized 0--1 rubric scale \\
Mean signed difference & \(\approx0.00\), indicating no systematic over- or under-scoring in this sample \\
\bottomrule
\end{tabularx}
\caption{\textbf{Expert validation of the LLM judge panel.} The stratified 60-response check validates the three-judge rubric mean against adjudicated expert scores. The result supports using the non-contestant judge panel as a practical scoring proxy, while the paper treats it as a sample validation rather than a full replacement for exhaustive human rescoring.}
\label{tab:human-judge-validation}
\end{table}

\paragraph{Rubric-only versus evidence-visible judging.}
Production scoring is rubric-conditioned grading with source-grounded rubrics: the judge sees the question, criterion-level rubric, and model response, not the full source or evidence windows. Source verification is front-loaded into item construction, D1/D7/D9 audits, and external acceptance review. As a spot check, we reran 124 stratified answer cells with Claude-Sonnet-4.6 under two conditions: the production rubric-only prompt and the same prompt with gold facts plus verbatim clue passages. The evidence-visible condition was slightly more generous on average (\(+3.6\) percentage points), and 6\% of cells moved down by 20 points or more. This check did not show broad rubric-only over-credit in the sampled cells.

\begin{table}[H]
\centering
\footnotesize
\setlength{\tabcolsep}{4pt}
\renewcommand{\arraystretch}{1.12}
\begin{tabularx}{\textwidth}{@{}L{0.16\textwidth}rrrrL{0.13\textwidth}Y@{}}
\toprule
\textbf{Check} & \textbf{\(n\)} & \textbf{Rubric-only} & \textbf{Evidence-visible} & \textbf{\(\Delta\)} & \textbf{Cells down \(\geq20\) pp} & \textbf{Read} \\
\midrule
Sonnet spot check & 124 & 65.8 & 69.4 & +3.6 pp & 6\% & Adding gold facts and clue passages does not reveal systematic over-credit by the production rubric-only judge; the evidence-visible condition is slightly more generous on average. \\
\bottomrule
\end{tabularx}
\caption{\textbf{Rubric-only versus evidence-visible judging.} Production scoring grades the question, criterion rubric, and model response; the spot-check condition additionally includes gold facts and verbatim clue passages. Small average shifts suggest that the source-grounded rubric carries most evidence constraints in this sample, while large downward moves flag responses that looked plausible under the rubric but contradicted or fabricated against the shown evidence.}
\label{tab:judge-evidence-visibility}
\end{table}

\paragraph{Observed failure modes.}
The most common rejected candidates were not low-quality questions in a generic sense. They were specifically invalid for source-internal multi-hop reasoning: a single local clue already determined the answer, a key fact was paraphrased beyond source support, a causal or abductive bridge was plausible but not evidenced, the item label mismatched its geometry, or the answer could be recovered without the document. These observations inform the release gates above rather than serving as an alternate benchmark result.

\subsection{Executable Reproducibility Protocol}
\label{app:reproducibility-protocol}

The public release package at \releaseRepo{} contains the locked task JSON, questions-only JSONL, source files, schema, checksums, Croissant/Responsible-AI metadata, evaluation scripts, reference scores, multi-judge score matrices, and a full evaluation protocol. The model-under-test prompt is instantiated without a system message, few-shot example, chain-of-thought request, evidence span, clue count, reference answer, or rubric:

\begingroup
\setlength{\fboxsep}{4pt}
\noindent\fbox{\begin{minipage}{0.96\linewidth}
\footnotesize
\textbf{Model-under-test prompt template.}\par\smallskip
\ttfamily
Answer using ONLY the document below. Include every specific detail from the text.\par
\smallskip
Question:\par
\{question\}\par
\smallskip
Document:\par
\{full\_document\}
\end{minipage}}
\endgroup
\medskip

The reported leaderboard uses this question-before-document ordering (Q-before-D) because it makes the task goal explicit before very long inputs are serialized. This is the standardized released route, not a claim that prompt order is irrelevant. The instruction ``Include every specific detail'' was intended to reduce omission of answer-critical facts; it can also encourage verbosity. Rubric scoring therefore credits only answer-critical criteria.

For models that accept decoding parameters, generation uses low-temperature decoding and a fixed output budget under the released harness. Benchmark scores are computed from logged model responses rather than from construction artifacts. Because closed endpoints can change over time, new evaluations should archive the endpoint snapshot and decoding configuration used for that run.

The released evaluation script is the source of truth for prompt serialization, input budgeting, status logging, and score normalization. New model runs should report their configuration separately from the locked benchmark release rather than editing the task artifact.

The L0--L7 context tiers are dataset strata based on the release field \texttt{doc\_token\_estimate}; they are used to balance and analyze source lengths. Endpoint-specific input handling is implemented in the evaluation script and archived with each run.

At a protocol level, each non-contestant judge receives the public question, criterion-level rubric, and model response, then returns criterion points and a total score. Scores are normalized to \([0,1]\), and the task score is the simple mean of the non-contestant judges for that logged response. The release includes both the averaged score matrix and per-judge matrices so that reviewers can inspect judge sensitivity without rerunning all model endpoints.

\subsection{Evaluation Boundary and Role Separation}
\label{app:evaluation-boundary}
\label{app:role-separation}

\paragraph{Evaluation boundary.}
The primary task is full-document and evidence-withheld. During evaluation, a model receives the full source and the question, but not evidence spans, clue counts, graph paths, reference answers, rubrics, validation notes, or manual-review artifacts. Clue-window prompts are reserved for construction diagnostics, answerability checks, and access diagnostics; they are not merged into the primary full-document leaderboard.

\begin{table}[!htbp]
\centering
\small
\setlength{\tabcolsep}{5pt}
\renewcommand{\arraystretch}{1.12}
\begin{tabularx}{\linewidth}{@{}L{0.30\linewidth}Y@{}}
\toprule
\textbf{Component} & \textbf{\systemName{} protocol} \\
\midrule
Model sees & Full source and public question. \\
Model does not see &
Evidence spans, clue counts, source-window hints, graph path, reference answer,
rubric, validation notes, or manual-review artifacts. \\
Output & Open-form answer. \\
Scoring & Criterion-level rubric scoring by three non-contestant judge models. \\
Main metric & Mean rubric credit from the frozen full-document score matrix. \\
Run metadata & Endpoint configuration is archived separately from the locked task artifact. \\
\bottomrule
\end{tabularx}
\caption{\textbf{Full-document evaluation boundary.}
Evaluated models receive only the full source and public question. Construction
and scoring artifacts remain hidden during evaluation.}
\label{tab:evaluation-boundary}
\end{table}

\paragraph{Benchmark release and evaluation logs.}
The final dataset contains 481 locked tasks. The main score panel reports the frozen three-judge rubric mean under the released full-document evaluation harness. Endpoint logs are retained as run provenance and do not redefine the locked task set or headline score panel.

\clearpage
\section{Additional Results and Diagnostic Ablations}
\label{app:additional-results}

The main text reports the benchmark-release leaderboard, context-tier curves, and evidence-geometry profile. This appendix keeps only supporting diagnostics that are needed to interpret those results: geometry-by-tier interaction, source-family slices, and an additive factor diagnostic. Unless otherwise noted, these slices are recomputed from the final benchmark-release multi-judge score matrix and should be read as diagnostics, not alternate leaderboards.

\begin{table*}[t]
\centering
\definecolor{wtA}{HTML}{D9EAD3}
\definecolor{wtB}{HTML}{EAF3F8}
\definecolor{wtC}{HTML}{FFF2CC}
\definecolor{wtD}{HTML}{FCE4D6}
\newcommand{\hi}[1]{\cellcolor{wtA}\textbf{#1}}
\newcommand{\good}[1]{\cellcolor{wtB}#1}
\newcommand{\midv}[1]{\cellcolor{wtC}#1}
\newcommand{\lowv}[1]{\cellcolor{wtD}#1}
\scriptsize
\setlength{\tabcolsep}{3.4pt}
\renewcommand{\arraystretch}{1.12}
\begin{tabular}{lrrrrrrr}
\toprule
\textbf{Geometry} & \textbf{L0} & \textbf{L1} & \textbf{L2} & \textbf{L3} & \textbf{L4} & \textbf{L5} & \textbf{L6} \\
\midrule
Forward & \hi{82.7 (12)} & \good{70.8 (8)} & \good{72.3 (9)} & \good{74.7 (9)} & \midv{62.9 (10)} & \lowv{51.0 (8)} & \good{72.0 (16)} \\
Intersection & \hi{87.5 (12)} & \hi{80.4 (10)} & \hi{85.9 (8)} & \hi{92.5 (8)} & \hi{92.7 (8)} & \hi{86.4 (10)} & \good{73.8 (13)} \\
Comparative & \hi{91.1 (12)} & \midv{62.2 (9)} & \hi{82.6 (9)} & \hi{88.4 (9)} & \hi{83.2 (8)} & \lowv{53.9 (8)} & \good{71.3 (12)} \\
Temporal & \hi{82.2 (13)} & \good{78.1 (8)} & \good{76.2 (10)} & \lowv{50.7 (8)} & \midv{64.4 (9)} & \lowv{46.2 (8)} & \good{70.2 (15)} \\
Abductive & \hi{88.6 (12)} & \lowv{56.7 (8)} & \midv{63.6 (9)} & \lowv{59.7 (8)} & \midv{65.1 (8)} & \midv{69.0 (8)} & \good{70.8 (12)} \\
Causal & \good{79.9 (15)} & \good{72.8 (8)} & \lowv{44.6 (8)} & \lowv{54.7 (8)} & \good{72.7 (8)} & \midv{61.0 (8)} & \good{75.7 (12)} \\
Counterfactual & \hi{82.4 (14)} & \midv{67.3 (8)} & \midv{62.3 (8)} & \lowv{44.3 (8)} & \midv{60.4 (8)} & \lowv{56.8 (8)} & \midv{62.2 (13)} \\
\bottomrule
\end{tabular}
\vspace{2pt}
\begin{minipage}{0.94\linewidth}
\centering\footnotesize
Entries are mean scores for the top-six headline systems, with item counts in parentheses. L7 is omitted because it has only three stress-probe tasks.
\end{minipage}
\caption{\textbf{Evidence geometry and context tier interact.} Entries are top-six mean scores with item counts in parentheses, grouped by primary geometry and document-length tier. The pattern is not a pure length law: some geometries are brittle even at mid lengths, while some long-tier cells remain robust for frontier systems.}
\label{tab:geometry-context}
\end{table*}

\begin{table*}[t]
\centering
\definecolor{wtWin}{HTML}{D9EAD3}
\definecolor{wtWeak}{HTML}{F4CCCC}
\scriptsize
\setlength{\tabcolsep}{3.2pt}
\renewcommand{\arraystretch}{1.10}
\begin{tabular}{lrrrrrrrr}
\toprule
\textbf{Source family} & \textbf{Gemini-3.1-Pro} & \textbf{Opus 4.8} & \textbf{GLM-5.2} & \textbf{GPT-5.5} & \textbf{Qwen3.7-Max} & \textbf{Opus 4.6} & \textbf{Gemini-2.5} & \textbf{MiniMax-M2.7} \\
\midrule
Technical reports & \cellcolor{wtWin}\textbf{77.7} & 74.3 & 70.9 & 68.9 & 75.4 & 70.1 & 64.1 & \cellcolor{wtWeak}36.1 \\
EN literature & \cellcolor{wtWin}\textbf{74.7} & 71.6 & 71.5 & 71.5 & 68.4 & 69.5 & 60.2 & \cellcolor{wtWeak}25.2 \\
CJK literature & 72.5 & \cellcolor{wtWin}\textbf{75.2} & 69.7 & 74.4 & 68.2 & 63.1 & 62.5 & \cellcolor{wtWeak}36.9 \\
\bottomrule
\end{tabular}
\vspace{2pt}
\begin{minipage}{0.94\linewidth}
\centering\footnotesize
Scores are diagnostic three-judge means within each source family under the same released scoring protocol.
\end{minipage}
\caption{\textbf{Source-family slice.} Scores are diagnostic benchmark-release means within technical reports, English literature, and the smaller CJK literature slice. The columns should be read as domain diagnostics rather than a multilingual generalization claim: different source families stress different evidence regimes, and no model dominates every family.}
\label{tab:source-type-ablation}
\end{table*}

\FloatBarrier
\begin{table}[H]
\centering
\footnotesize
\setlength{\tabcolsep}{5pt}
\renewcommand{\arraystretch}{1.06}
\begin{tabular}{lrr}
\toprule
\textbf{Factor group} & \textbf{Single-group \(R^2\)} & \textbf{Incremental \(R^2\)} \\
\midrule
Model identity & 0.189 & 0.189 \\
Context tier & 0.130 & 0.032 \\
Evidence geometry & 0.022 & 0.028 \\
Structural metadata & 0.104 & 0.005 \\
Source family & 0.002 & 0.001 \\
Full additive model & 0.353 & -- \\
\bottomrule
\end{tabular}
\caption{\textbf{ANOVA-style diagnostic over benchmark-release answer cells.} The additive model includes model identity, tier, geometry, source family, and structural metadata; incremental \(R^2\) is the drop from removing a group. This is descriptive rather than causal: length effects coexist with model, geometry, source-family, and structural effects.}
\label{tab:factor-diagnostic}
\end{table}


\clearpage

\newpage

\end{document}